\documentclass[10pt,twocolumn,letterpaper]{article}

\usepackage{cvpr}              

\usepackage{algpseudocode}
\usepackage{subcaption}
\usepackage{amsmath} 
\usepackage{amssymb}
\usepackage{tikz}
\usepackage{mathtools}
\usepackage{booktabs}
\usepackage{array}
\usepackage{makecell}
\usepackage{lipsum}
\usepackage{colortbl}
\usepackage{multirow}
\usepackage{arydshln}
\usepackage{color, colortbl}
\definecolor{LightCyan}{rgb}{0.88,1,1}
\usepackage{adjustbox}
\newcolumntype{C}[1]{>{\centering\arraybackslash}p{#1}}
\usepackage{pifont}

\definecolor{myLightGray}{RGB}{220, 220, 220}
\definecolor{apricot}{RGB}{251, 206, 177}   
\definecolor{lightblue}{RGB}{173, 216, 230} 
\newcommand\blfootnote[1]{%
\begingroup
\renewcommand\thefootnote{}\footnote{#1}%
\addtocounter{footnote}{-1}%
\endgroup
}
\definecolor{cvprblue}{rgb}{0.21,0.49,0.74}
\usepackage[pagebackref,breaklinks,colorlinks,allcolors=cvprblue]{hyperref}

\def\paperID{5268} 
\def\confName{CVPR}
\def\confYear{2025}

\title{Towards Generalizable Scene Change Detection}

\author{Jae-Woo Kim\textsuperscript{\rm 1} and Ue-Hwan Kim\textsuperscript{\rm 1}\thanks{Corresponding author}\\
\textsuperscript{\rm 1}Department of AI Convergence\\
Gwangju Institute of Science and Technology, Gwangju, South Korea \\
{\tt\small kjw01124@gm.gist.ac.kr, uehwan@gist.ac.kr}
}

\makeatletter
\@ifundefined{LN@col}{%
  \def\@LN@col#1{}%
}{}
\@ifundefined{LN}{%
  \def\@LN#1#2{}%
}{}
\makeatother

\begin{document}
\twocolumn[%
    \vspace{-1.3cm}
    \maketitle
    \vspace{-0.5cm}
    \begin{center}
        \footnotesize
        \setlength{\tabcolsep}{1pt} 
        \renewcommand{\arraystretch}{1.2} 

        \begin{tabular}{>{\centering\arraybackslash}m{17cm}}
            \adjustbox{valign=t}{\includegraphics[width=\linewidth]{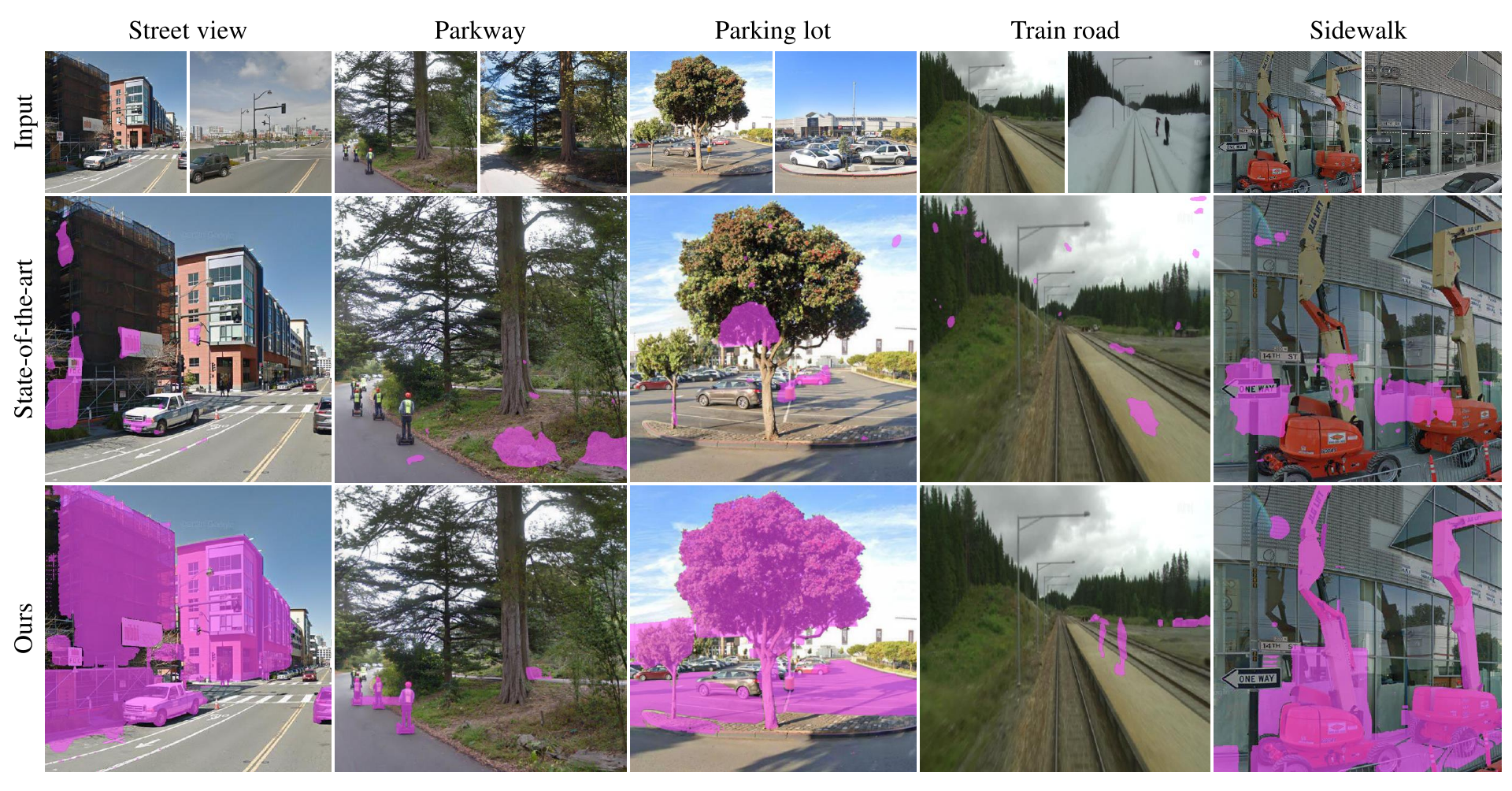}}
        \end{tabular}

        \captionof{figure}{\textbf{Comparative results of the current state-of-the-art model and our GeSCF under various unseen environments (ChangeVPR).} GeSCF outperforms with precise boundaries and edges, where the state-of-the-art model hardly captures changes.}
        \label{fig:title}
    \end{center}
    \vspace{0.2cm}
]

\blfootnote{Code and dataset are available \href{https://github.com/1124jaewookim/towards-generalizable-scene-change-detection}{here}.}
\blfootnote{* Corresponding author}

\begin{abstract}

While current state-of-the-art Scene Change Detection (SCD) approaches achieve impressive results in well-trained research data, they become unreliable under unseen environments and different temporal conditions; in-domain performance drops from 77.6\% to 8.0\%  in a previously unseen environment and to 4.6\%  under a different temporal condition---calling for generalizable SCD and benchmark. In this work, we propose the Generalizable Scene Change Detection Framework (GeSCF), which addresses unseen domain performance and temporal consistency---to meet the growing demand for anything SCD. Our method leverages the pre-trained Segment Anything Model (SAM) in a zero-shot manner. For this, we design Initial Pseudo-mask Generation and Geometric-Semantic Mask Matching---seamlessly turning user-guided prompt and single-image based segmentation into scene change detection for a pair of inputs without guidance. Furthermore, we define the Generalizable Scene Change Detection (GeSCD) benchmark along with novel metrics and an evaluation protocol to facilitate SCD research in generalizability. In the process, we introduce the ChangeVPR dataset, a collection of challenging image pairs with diverse environmental scenarios---including urban, suburban, and rural settings. Extensive experiments across various datasets demonstrate that GeSCF achieves an average performance gain of 19.2\% on existing SCD datasets and 30.0\% on the ChangeVPR dataset, nearly doubling the prior art performance. We believe our work can lay a solid foundation for robust and generalizable SCD research.
\end{abstract}
    
\section{Introduction}
\label{sec:intro}

Scene Change Detection (SCD) \cite{Radke2005ImageCD} is a pivotal technology that enables a wide range of applications: visual surveillance \cite{wu2020learn}, anomaly detection \cite{Huang2022RegistrationBF}, mobile robotics \cite{Nehmzow2000MobileRA}, and autonomous vehicles \cite{Janai2017ComputerVF}. The ability to accurately identify meaningful changes in a scene across different time steps---despite challenges such as illumination variations, seasonal changes, and weather conditions---plays a key role in the system's effectiveness and reliability.

Recent SCD models have achieved remarkable improvement by leveraging deep features \cite{Alcantarilla2016StreetviewCD, Varghese2018ChangeNetAD} and advancing model architectures \cite{Sakurada2018WeaklySS, Chen2021DRTANetDR, Park2022DualTL, wang2023reduce}. However, this progress raises a fundamental question: \textbf{\textit{``Can these models detect arbitrary real-world changes beyond the scope of research data?”}} Our findings, as shown in \cref{fig:title}, indicate that their reported effectiveness does not hold in real-world applications. Specifically, we observe that they produce inconsistent change masks when the input order is reversed and exhibit significant performance drops when deployed to unseen domains with different visual features. It is because current SCD methods rely heavily on their training datasets, which are often limited in size \cite{Alcantarilla2016StreetviewCD, Sakurada2015ChangeDF}, sparse in coverage \cite{Sakurada2018WeaklySS, Alcantarilla2016StreetviewCD, Sakurada2015ChangeDF}, and predominantly synthetic \cite{Park2021ChangeSimTE, Park2022DualTL, Lee2024SemiSupervisedSC} due to the costly change annotation.

To address these challenges, we introduce the \textit{\textbf{Generalizable Scene Change Detection Framework (GeSCF)}}, the first zero-shot scene change detection method that operates robustly regardless of the temporal input order and environmental conditions. Our approach builds upon the Segment Anything Model (SAM) \cite{Kirillov2023SegmentA}, a pioneering vision foundation model for image segmentation. While SAM excels at segmenting \textit{anything} within a single image, directing SAM to identify and segment \textit{changes} between two input images presents a significant challenge. This difficulty arises because SAM is designed for promptable interactive segmentation, relying on user-guided prompts and single-image inputs, whereas scene change detection necessitates processing image pairs to identify changes. To bridge this gap, we propose two innovations: the Initial Pseudo-mask Generation and the Geometric-Semantic Mask Matching. By analyzing the localized semantics of the SAM’s feature space, we effectively binarize pixel-level change candidates with zero additional cost. Additionally, we leverage the geometric properties of SAM's class-agnostic masks and the semantics of mask embeddings to refine the final change masks---incorporating object-level information as well. 

Furthermore, we introduce the Generalizable Scene Change Detection (GeSCD) benchmark by developing novel metrics and an evaluation protocol to foster SCD research in generalizability; most conventional SCD methods have focused on individual benchmarks separately rather than the generalizability to unseen domains and temporal consistency. We believe our GeSCD can meet the growing needs for developing \textit{anything} SCD in the era of diverse \textit{anything} models \cite{Kirillov2023SegmentA,Zou2023SegmentEE,Yang2023TrackAS,Wang2023DetectingEI,Yu2023InpaintAS,Xie2023EditEA} with strong zero-shot capabilities. Concretely, our GeSCD performs extensive cross-domain evaluations to rigorously test the generalizability across diverse environments and assesses temporal consistency quantitatively. Our dual-focused evaluation strategy not only ensures the robustness and reliability of a method but also sets a new benchmark in the SCD field. In the process of designing GeSCD, we collect the ChangeVPR dataset, which comprises carefully annotated images sourced from three prominent Visual Place Recognition (VPR) datasets. This comprehensive dataset includes urban, suburban, and rural environments under challenging conditions, significantly expanding the traditional scope of SCD domains.

To summarize, our contributions are as follows: 

\begin{enumerate}
    \item \textbf{Problem Formulation.} We introduce GeSCD, a novel task formulation in scene change detection. To the best of our knowledge, this is the first comprehensive effort to address the generalization problem and temporal consistency in SCD research.
    
    \item \textbf{Model Design.} To tackle the GeSCD task, we propose GeSCF, the first zero-shot scene change detection model. GeSCF exhibits complete temporal consistency and demonstrates strong generalizability over previous SCD models that are tightly coupled to their training datasets---ours resulting in a substantial performance gain in unseen domains.
    
    \item \textbf{Benchmark Set up.} We present new evaluation metrics, the ChangeVPR dataset, and an evaluation protocol that effectively measures an SCD model's generalizability. These contributions provide a solid foundation to guide and inspire future research in the field.
\end{enumerate}
\section{Related Works}

\textbf{Segment Anything Model.} Segment Anything Model (SAM) \cite{Kirillov2023SegmentA} has set a new standard in image segmentation and made significant strides in various computer vision domains: medical imaging \cite{Zhang2023InputAW}, camouflaged object detection \cite{Tang2023CanSS}, salient object detection \cite{Ma2023SegmentAI}, image restoration \cite{Yu2023InpaintAS}, image editing \cite{Xie2023EditEA}, and video object tracking \cite{Yang2023TrackAS}. By leveraging geometric prompts---such as points or bounding boxes---SAM showcases exceptional zero-shot transfer capabilities across diverse segmentation tasks and unseen image distributions. While SAM demonstrates impressive abilities, its potential for zero-shot scene change detection remains largely unexplored. In this work, we extend SAM's utility beyond single-image segmentation by introducing a novel, training-free approach that guides SAM to detect changes between a pair of natural images.

\noindent\textbf{Change Detection.} In the field of Change Detection (CD), the research falls broadly into three areas based on data characteristics: remote sensing CD, video sequence CD, and natural scene CD---the focus of our work. Remote sensing CD \cite{Saha2022SupervisedCD, Chen2022SARASNetSA,Bernhard2023MapFormerBC,Hao2023SemisupervisedLP,Bandara2022RevisitingCR,Bergamasco2022UnsupervisedCD,Noh2022UnsupervisedCD} involves detecting surface changes over time using data captured by satellite or aerial platforms, providing a high-altitude perspective of phenomena such as urbanization, deforestation, and disaster damage. Additionally, video sequence CD focuses on segmenting frames into foreground and background regions, usually corresponding to moving objects \cite{Mandal20203DCDSI, Akilan2020A3C}. In contrast to these CDs, natural scene CD aims to detect localized changes from a ground-level perspective such as movement of vehicles \cite{Alcantarilla2016StreetviewCD}, pedestrians \cite{Sakurada2018WeaklySS}, the appearance and disappearance of objects \cite{Alcantarilla2016StreetviewCD, Sakurada2018WeaklySS, Sakurada2015ChangeDF}, and significant background changes like the construction or demolition of buildings \cite{Sakurada2015ChangeDF, Sakurada2018WeaklySS}. Moreover, the task inherently involves misaligned and noisy images due to the nature of data acquisition, as images are often captured by cameras mounted on moving vehicles or robots \cite{Li2024UMADUO}. Overall, our work concentrates natural scene CD \cite{wang2023reduce, Park2022DualTL, Alcantarilla2016StreetviewCD, Chen2021DRTANetDR, Sakurada2017DenseOF, Sakurada2018WeaklySS, Varghese2018ChangeNetAD, Lee2024SemiSupervisedSC}; for the remainder of this paper, we will refer to natural Scene CD as SCD.

\noindent\textbf{Scene Change Detection (SCD).} In exsiting SCD benchmarks, most methods are supervised \cite{Sakurada2018WeaklySS, Chen2021DRTANetDR, wang2023reduce, Park2022DualTL,  Alcantarilla2016StreetviewCD, Sakurada2017DenseOF,Varghese2018ChangeNetAD} or semi-supervised \cite{Lee2024SemiSupervisedSC}, heavily optimized and evaluated on specific training datasets---leading to low generalizability. Although several works have proposed self-supervised pre-training strategies \cite{Ramkumar2022DifferencingBS} or leveraged temporal symmetry \cite{wang2023reduce}, they still exhibit significant performance gaps when deployed to unseen data. Moreover, the symmetric structure relies on a specific prior knowledge of the domain, rendering it impractical for unknown domains without the proper inductive bias. In contrast, our GeSCF stands as a unified, training-free framework displaying robust performance on unseen data while preserving symmetric architecture for all settings.

\noindent\textbf{Segment Anything with Change Detection.} Previous works have primarily leveraged SAM in the remote sensing CD with the Parameter-Efficient Fine-Tuning (PEFT) strategy \cite{Xu2023ParameterEfficientFM}. For instance, several works leveraged SAM variants \cite{Zhao2023FastSA,Zhang2023FasterSA} with learnable adaptors \cite{Xu2023ParameterEfficientFM}, fine-tuning adaptor networks and change decoders tailored to specific datasets \cite{Ding2023AdaptingSA,Mei2024SCDSAMAS}.  In contrast to these approaches, our method is the first solid SAM-integrated framework in SCD, leveraging SAM's internal byproducts to effectively binarize change candidates without any guidance and learnable parameters. Moreover, we further exploit valuable prior information \cite{Yu2023InpaintAS,Xie2023EditEA,Ahmadi2023ApplicationOS,Giannakis2023DeepLU,Chen2023SegmentAM} from SAM's class-agnostic masks---enabling robust zero-shot SCD across a broad range of domains for the first time.

\begin{figure*}[t]
  \centering
  \includegraphics[width=0.85\linewidth]{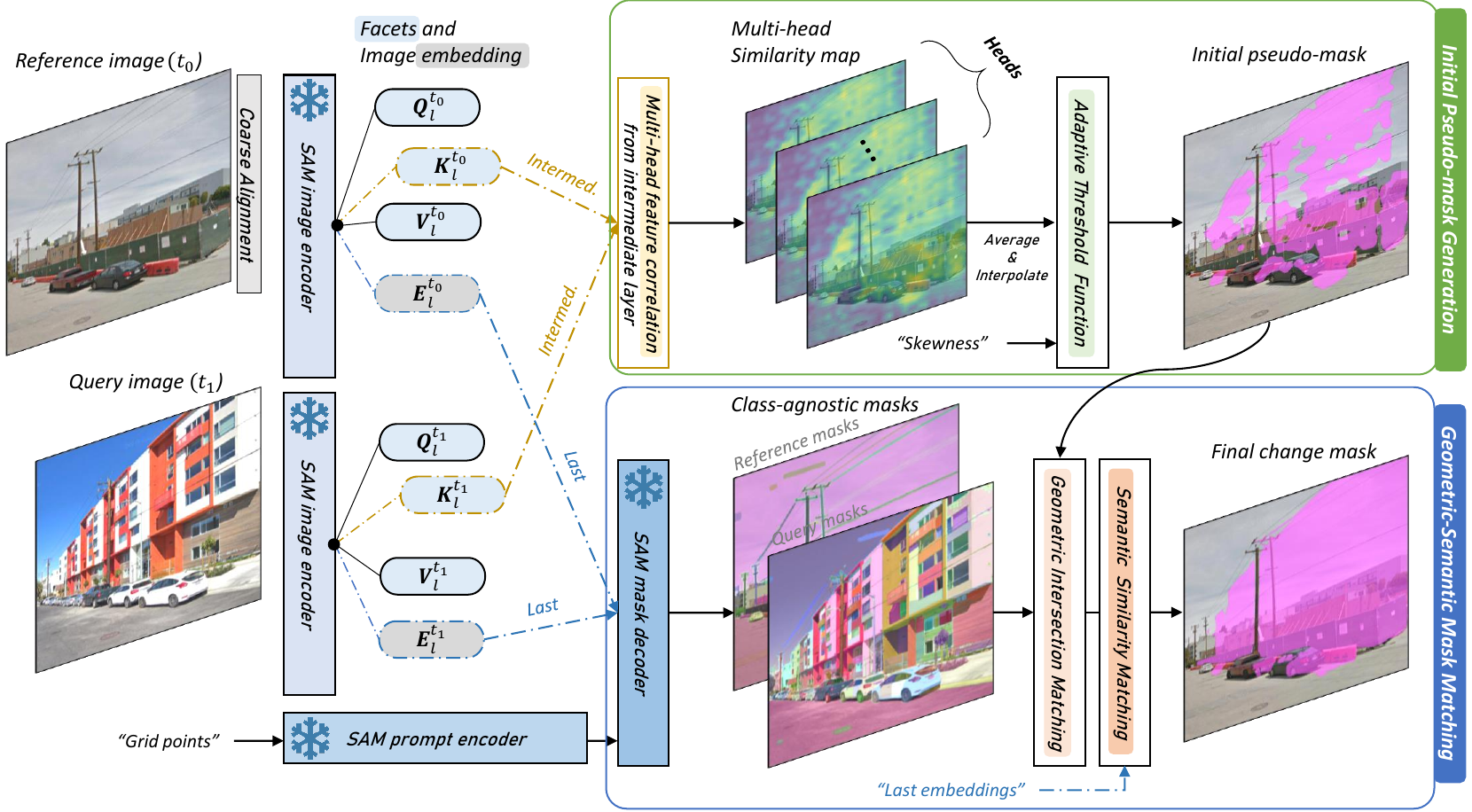}
  \caption{\textbf{Illustration of the proposed GeSCF pipeline.} The GeSCF pipeline consists of two major steps: (1) initial pseudo-mask generation and (2) geometric-semantic mask matching. First, we intercept facet features from the SAM image encoder and correlate them to obtain multi-head similarity maps, which are then converted into pseudo-masks using an adaptive threshold function. Next, SAM's class-agnostic masks and last image embeddings are utilized to refine these pseudo-masks based on both geometric and semantic information.}
  \label{fig:overview}

\end{figure*}

\section{GeSCF}
\subsection{Motivation and Overview}
Despite the abundance of web-scale data \cite{Radford2021LearningTV, Oquab2023DINOv2LR, Berton2022RethinkingVG} and the advent of various zero-shot generalizable models \cite{Kirillov2023SegmentA,Yang2024DepthAU,Ke2023RepurposingDI}, current SCD still suffers from the curse of datasets due to the costly nature of change annotation \cite{Zheng2021ChangeIE}. Therefore, our research motivation arises from \textit{\textbf{how SCD can benefit from recent web-scale trained models}} like SAM.  By addressing this question, we aim to overcome the longstanding obstacle of creating a generalizable SCD model---culminating in our proposed GeSCF model.

\cref{fig:overview} gives an overview of the GeSCF pipeline. GeSCF handles the technical gap between SAM designed for promptable interactive segmentation with single-image inputs and SCD for identifying changes with image pairs through two key stages: Initial Pseudo-mask Generation and Geometric-Semantic Mask Matching. First, we intercept and correlate feature facets---one of query, key, and value---from the image encoder to obtain rich, multi-head similarity maps; then, we transform these similarity maps into a binary pseudo-mask by adaptively thresholding the low-similarity pixels using a \textit{skewness}-based algorithm. Finally, we elaborate the pseudo-mask by leveraging the geometric properties of SAM's class-agnostic masks; then, we further validate these masks by comparing the semantic similarities of the corresponding mask embeddings between bi-temporal images, ensuring that the detected changes are meaningful and contextually accurate.



\subsection{Preliminary}
Since our GeSCF exploits a set of image features intercepted from the SAM image encoder, we first
recap how such features are obtained.

\noindent\textbf{Feature Facets.} The SAM image encoder employs a Vision Transformer (ViT) architecture \cite{Dosovitskiy2020AnII}, consisting of multi-head self-attention layers and multi-layer perceptrons within each ViT block \cite{Dosovitskiy2020AnII, Vaswani2017AttentionIA}. In the multi-head self-attention layer of the $l$-th ViT block, the query, key, and value facets are denoted by $\mathbf{QKV}_l \in \mathbb{R}^{3 \times N \times H \times W \times C}$, where $N$, $H$, $W$, and $C$ represent the number of heads, height, width, and channel dimensions of facets, respectively. 


\noindent\textbf{Image Embedding and Mask Embedding.} Similarly, we extract the image embedding $\mathbf{E}_l \in \mathbb{R}^{H \times W \times C}$ from the final multi-layer perceptron layer of the $l$-th ViT block. Furthermore, given the image embedding $\mathbf{E}_l$ and an arbitrary binary mask $\overline{\mathbf{m}}$, we compute the mask embedding $\mathcal{M}_l$ by averaging image embedding $\mathbf{E}_l$ over all spatial positions where the binary mask $\overline{\mathbf{m}}$ is non-zero, obtaining a single vector representation of the masked image.

\subsection{Initial Pseudo-mask Generation} 
As demonstrated in \cite{Kirillov2023SegmentA}, SAM preserves semantic similarities among mask embeddings within the same natural scene. Furthermore, as observed in prior studies \cite{Amir2021DeepVF, Caron2021EmergingPI}, attention maps can capture semantically meaningful objects in images. Building on these foundational insights, we expand SAM's feature space utilization by extending its application to bi-temporal images and leveraging multi-head feature facets from various layers rather than relying solely on single-image embeddings \cite{Kirillov2023SegmentA}.

\noindent\textbf{Multi-head Feature Correlation.} Given a bi-temporal RGB image pair, we first estimate a coarse transformation \cite{Li2024UMADUO, Truong2021PDCNetEP, Koguciuk2021PerceptualLF} relating the bi-temporal images. Then, we intercept the feature facets $\mathbf{F}_{l,n}^{t_0}$,  $\mathbf{F}_{l,n}^{t_1}$ (one of $\mathbf{Q}$, $\mathbf{K}$, and $\mathbf{V}$) for the $n$-th head from the $l$-th ViT block and compute multi-head feature correlation as follows:

\begin{equation}
    \overline{\mathbf{S}}_{l,n}^{t_0 \leftrightarrow t_1} = \mathbf{F}_{l,n}^{t_0}  \prescript{\top}{}{\mathbf{F}_{l,n}^{t_1}}, 
    \label{eq:corr}
\end{equation}

\begin{equation}
    \mathbf{S}_{l}^{t_0 \leftrightarrow t_1} = \xi\left( \sum_{n=1}^{N} \overline{\mathbf{S}}_{l,n}^{t_0 \leftrightarrow t_1} \right) \vspace{0.1cm},
\end{equation}
\noindent where $\xi(\cdot)$ performs spatial reshaping and bilinear interpolation to the input image size. The correlation in \cref{eq:corr} is a commutative operation that generates identical similarity maps $\overline{\mathbf{S}}_{l,n}$ even with the reversed order. Subsequently, the similarity maps are averaged over the head dimension---leveraging the semantic diversity across the $N$ heads.

\begin{figure}[t]
  \centering
  \includegraphics[width=0.95\linewidth]{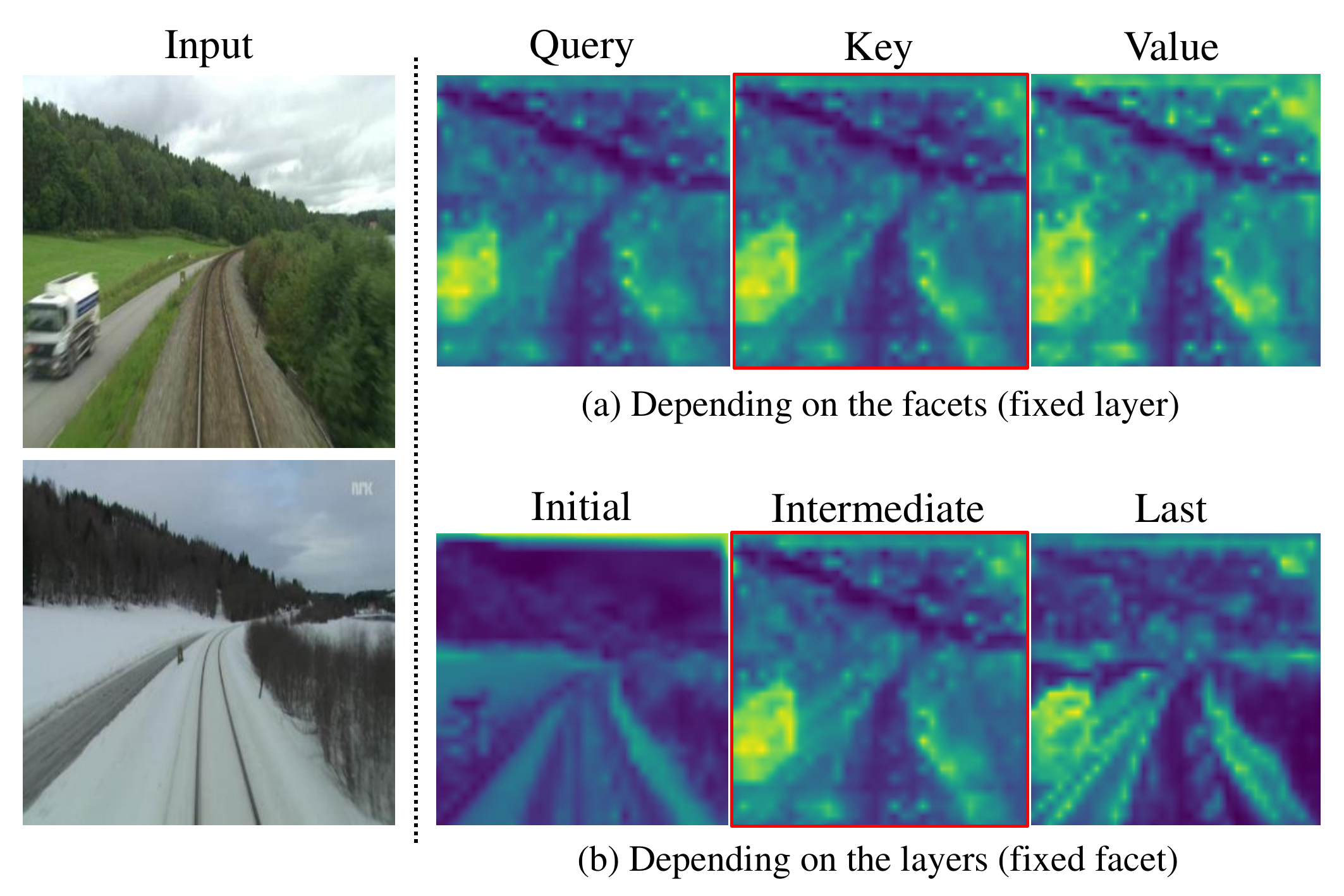}
  \caption{\textbf{Visualization of the similarity map depending on the facets and the layers.} We use key facets from the intermediate layer of the SAM ViT image encoder, which is highlighted with a red bounding box.}
  \label{fig:sim}
\end{figure}

\noindent\textbf{Facet and Layer Selection.} As shown in \cref{fig:sim}
, the similarity map of SAM's facets effectively highlights semantic changes while remaining relatively unaffected by visual variations, such as seasonal or illumination differences. Concretely, our feature selection strategy is guided by the following principles: (a) semantic changes should appear prominently in the similarity map compared to other visual variations; (b) the similarity values in changed regions should clearly contrast with surrounding areas; and (c) artifacts in unchanged regions should be minimized or appear faint. We empirically observe that all facets satisfy principle (a); however, principles (b) and (c) are particularly noticeable when using the key (or query) facet over the value facet. Moreover, we find that these principles hold more strongly in intermediate layers than in the initial or last layers. Consequently, we leverage the similarity map of key facets from intermediate layers as inputs to the subsequent steps in the pseudo-mask generation. For the detailed quantitative analysis, please refer to the supplementary material.

\noindent\textbf{Adaptive Threshold Function.} 
To create a binary mask from an arbitrary similarity map, the most straightforward approach is to apply a fixed threshold (e.g., $p=0.5$) \cite{Furukawa2020SelfsupervisedSA}, where similarity scores below this threshold are classified as changes. However, this fixed threshold approach is inherently limited, as it fails to account for the relative nature of ``\textit{change}" within each similarity map. For instance, a similarity score of 0.7 may signify a change if all other values are close to one, whereas it may not indicate a change if most values are near zero. Consequently, the perception of ``\textit{change}" is context-dependent, necessitating an adaptive thresholding method that accounts for the relative distribution of similarity scores within each map.

\begin{figure}[t]
    \centering
    \includegraphics[width=\linewidth]{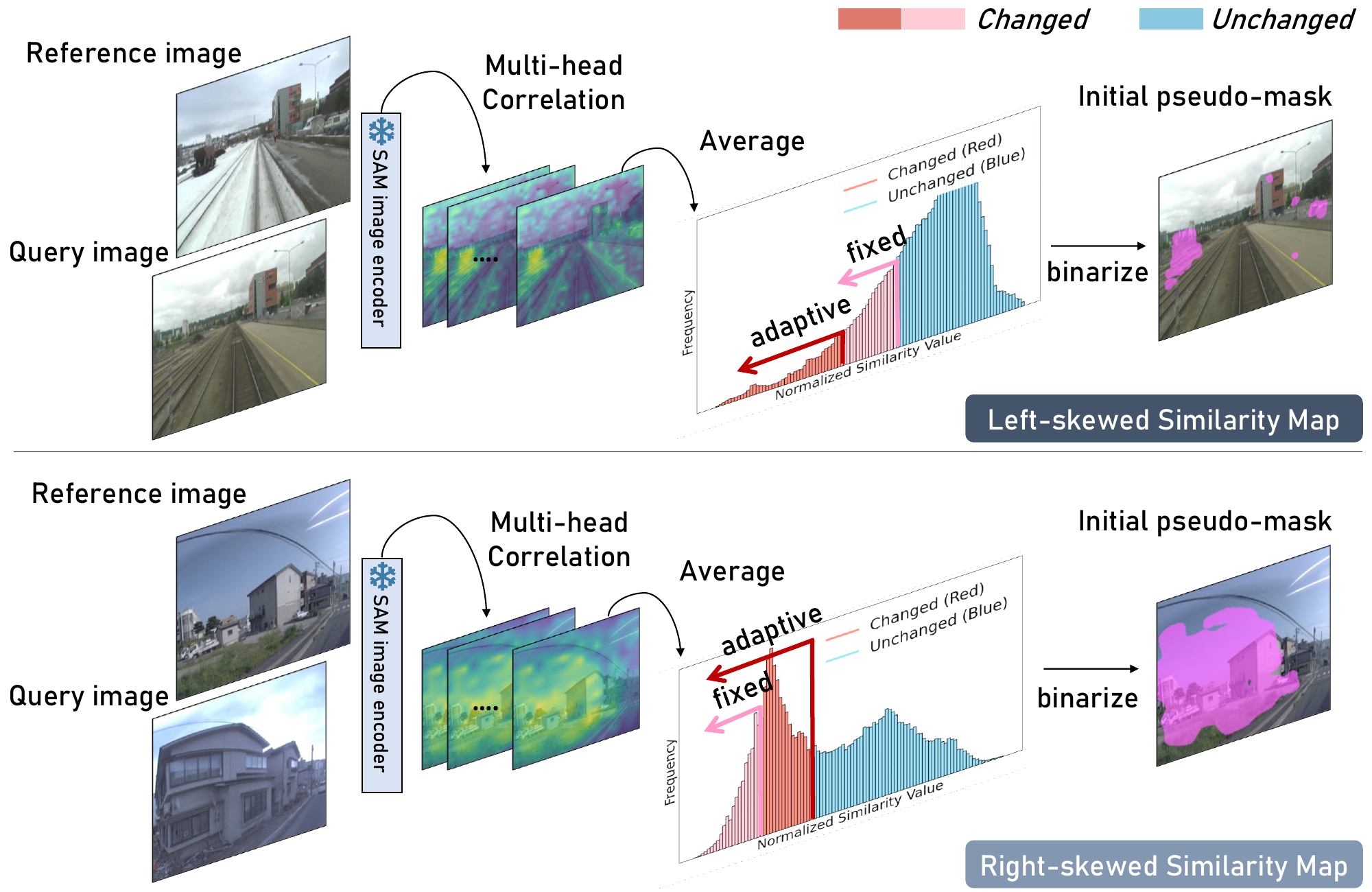}
    \caption{\textbf{Illustration of the adaptive thresholding process.} We dynamically adjust the threshold based on the skewness of the distribution to generate the initial pseudo-masks.}
    \label{fig:skw}
\end{figure}

Therefore, a critical factor to binarize the similarity map is the skewness ($\gamma$) \cite{Ramachandran2009MathematicalSW} of the similarity distribution (see \cref{fig:skw}). For right-skewed distributions, where the majority of pixels display lower similarity scores with a long tail of higher scores, a lower threshold is required to capture a larger portion of the distribution as changes. Conversely, for left-skewed distributions, where most scores are high with a small tail of lower values, a higher threshold is needed to avoid false positives. To this end, we propose an adaptive threshold function for dynamic adjustment based on the skewness of the distribution---enabling a more accurate and context-sensitive method for pseudo-mask generation, as detailed in the following formulation:

\begin{equation} \mathbf{F}(\gamma) = b_{\gamma} + c \cdot \operatorname{sign}(\gamma) \cdot s_{\gamma} \cdot \gamma, \end{equation}

\noindent where $b_{\gamma}$ is the baseline threshold, $c$ is a normalization constant, $s_{\gamma}$ is the skewness sensitivity factor, $\gamma$ is the skewness of the distribution, and $\operatorname{sign}(\cdot)$ adjusts the threshold direction based on the skewness. By applying the computed adaptive threshold to the similarity map---normalized using the mean absolute deviation (MAD) \cite{Ramachandran2009MathematicalSW}---we obtain the initial pseudo-mask essential for the subsequent Geometric-Semantic Mask Matching.

\subsection{Geometric-Semantic Mask Matching} 
Building upon the initial pseudo-masks, we elevate our focus to detecting object-level changes by using SAM's class-agnostic object proposals. This transition from pixel-wise analysis to object-level consideration enables a more comprehensive and interpretable change mask. Here, we introduce two matching strategies: geometric intersection matching (GIM) and semantic similarity matching (SSM).

\noindent\textbf{Geometric Intersection Matching.} 
The main idea of this strategy is to select SAM masks by evaluating their overlap with the initial pseudo-masks. We calculate the intersection ratio $\alpha$  between each SAM mask and the pseudo-mask, retaining only those masks with $\alpha$$>$$\alpha_{t}$, where $\alpha_t$ is a threshold. Since changes can occur bi-temporally, we apply the process for both $t_0$ and $t_1$ images---maintaining commutativity for the proposed GeSCF.

\noindent\textbf{Semantic Similarity Matching.} Although GIM provides reasonable masks for potential object-level changes, a set of unchanged regions is included due to noise in the initial pseudo-masks. To address this issue, we perform semantic verification of the overlapping regions by calculating the cosine similarity between the mask embeddings of the bi-temporal images. Specifically, for each overlapping mask $\overline{m}_{o}$, we extract the corresponding mask embeddings 
$\mathcal{M}_{l,o}^{t_0}$ and $\mathcal{M}_{l,o}^{t_1}$ from the $l$-th image embeddings $\mathbf{E}_l$ at times $t_0$ and $t_1$, respectively. We then compute a change confidence score using the cosine similarity $c(\mathcal{M}_{l,o}^{t_0}, \mathcal{M}_{l,o}^{t_1})$ which allows further refinement of the masks selected by GIM and generates the final change mask $\mathbf{Y}_{\text{pred}}$. Through a layer-wise analysis, we empirically observe that semantic differences are more pronounced in the last layer compared to the initial and intermediate layers, thereby utilizing final image embeddings for our SSM process\footnote{Please refer to the supplementary material for the analysis.}.

\section{GeSCD}
Since the introduction of the first SCD benchmark \cite{Sakurada2015ChangeDF} in 2015, the generalizability to unseen domains and temporal consistency of predictions have not been consistently or comprehensively addressed in the SCD field. Most traditional SCD approaches train and evaluate models on individual datasets separately \cite{wang2023reduce, Park2022DualTL, Alcantarilla2016StreetviewCD, Chen2021DRTANetDR, Sakurada2017DenseOF, Sakurada2018WeaklySS, Varghese2018ChangeNetAD, Lee2024SemiSupervisedSC}. Only \cite{Ramkumar2022DifferencingBS} performed cross-domain evaluation, however, it still suffers from domain gaps and limited training datasets. Moreover, in contrast to the remote sensing CD \cite{Zheng2021ChangeIE}, the temporal consistency is often overlooked in model design \cite{Park2022DualTL, Alcantarilla2016StreetviewCD, Chen2021DRTANetDR, Sakurada2017DenseOF, Sakurada2018WeaklySS, Varghese2018ChangeNetAD, Lee2024SemiSupervisedSC} or training objectives \cite{wang2023reduce, Park2022DualTL, Alcantarilla2016StreetviewCD, Chen2021DRTANetDR, Sakurada2017DenseOF, Sakurada2018WeaklySS, Varghese2018ChangeNetAD, Lee2024SemiSupervisedSC} in the SCD field. The temporal symmetry proposed in \cite{wang2023reduce} is unsuitable for real-world applications since it assumes perfect inductive bias of the application domain, which is not feasible in real-world scenarios. Further, as we are in the age of diverse \textit{anything} models \cite{Kirillov2023SegmentA,Zou2023SegmentEE,Yang2023TrackAS,Wang2023DetectingEI,Yu2023InpaintAS,Xie2023EditEA} with strong zero-shot prediction and generalizability, the necessity of \textit{anything} SCD model that can adapt to various change scenarios has become increasingly important in the research community. 

Based on these requirements, we propose \textit{GeSCD}---a novel task approach that addresses the generalizability of broader scenarios and temporal consistency of SCD models. By pioneering new metrics, an evaluation dataset, and a comprehensive evaluation protocol, our approach fulfills the critical need for SCD research that is genuinely applicable and effective across diverse settings.

\noindent\textbf{Metrics.} To evaluate the environmental and temporal robustness of the methods simultaneously, we report conventional metrics (e.g.,  Intersection over Union and F1-score) for both temporal directions in contrast to previous methods that typically report performance for a single temporal direction \cite{wang2023reduce, Park2022DualTL, Alcantarilla2016StreetviewCD, Chen2021DRTANetDR, Sakurada2017DenseOF, Sakurada2018WeaklySS, Varghese2018ChangeNetAD, Lee2024SemiSupervisedSC}. Furthermore, we propose the Temporal Consistency (TC) metric by measuring the union intersection between $t0$$\rightarrow$$t1$ and $t1$$\rightarrow$$t0$ predictions as follows:

\begin{equation}
    \text{Temporal Consistency (TC)} = \frac{\mathbf{Y}^{t0 \rightarrow t1}_{\text{pred}} \cap \mathbf{Y}^{t1 \rightarrow t0}_{\text{pred}}}{\mathbf{Y}^{t0 \rightarrow t1}_{\text{pred}} \cup \mathbf{Y}^{t1 \rightarrow t0}_{\text{pred}}},
\end{equation}

\noindent where the proposed TC score indicates how much the SCD algorithms can generate consistent change masks in bi-directional orders. 

\noindent\textbf{Evaluation Datasets.}
First, we consider three standard SCD datasets with different characteristics: VL-CMU-CD \cite{Alcantarilla2016StreetviewCD}, TSUNAMI \cite{Sakurada2015ChangeDF}, and ChangeSim \cite{Park2021ChangeSimTE}. These datasets represent urban environments in the USA, disaster-impacted urban areas in Japan, and industrial indoor settings within simulation environments, respectively. Furthermore, for the quantitative evaluation on broader unseen domains, we create a new dataset named the \textit{ChangeVPR}. The ChangeVPR comprises 529 image pairs collected from the SF-XL (urban, U) \cite{Berton2022RethinkingVG}, St Lucia (suburban, S) \cite{Milford2008MappingAS}, and Nordland (rural, R) \cite{Snderhauf2013AreWT} datasets, which are widely used in the Visual Place Recognition (VPR) research. We carefully sampled image pairs from each dataset to reflect various SCD challenges such as weather conditions and seasonal changes, and hand-labeled the ground-truth change masks for each image pair. We employ ChangeVPR only for the evaluation to assess unseen domain performance. Please refer to the supplementary material for more details.

\noindent\textbf{Evaluation Protocol.} We perform extensive cross-domain evaluation for SCD models. First, we train each model on the three standard SCD datasets; the training stage results in three distinctive models for each method. Then, we assess each of the three distinctive models in two stages. First, we evaluate the three models on the three conventional SCD datasets; this process totals nine assessments. If the method of interest does not involve training, we evaluate the model on the three datasets (three assessments). Next, we evaluate the models on the proposed ChangeVPR dataset. Since we use ChangeVPR only for evaluation, this stage contains three assessments per model (due to three splits of ChangeVPR). As a result, the proposed protocol estimates performance on both seen and unseen domains---ensuring a thorough evaluation of SCD performance in various environmental scenarios with arbitrary real-world changes.

\section{Experiments}
\begin{table*}[t]
    \footnotesize
    \centering
    \setlength{\tabcolsep}{6.2pt}
    \renewcommand{\arraystretch}{0.5} 
    \begin{tabular}{@{}>{\raggedright}m{3.5cm} @{}>{\centering\arraybackslash}m{2cm} *{10}{>{\centering\arraybackslash}m{0.75cm}}@{}}
        \toprule
        \multirow{3}{*}{\textbf{Method}} & \multirow{3}{*}{\textbf{Training}} & \multicolumn{3}{c}{VL-CMU-CD} & \multicolumn{3}{c}{TSUNAMI} & \multicolumn{3}{c}{ChangeSim} & \multirow{3}{*}{\textbf{Avg.}}  \\
        \cmidrule(lr){3-5} \cmidrule(lr){6-8} \cmidrule(lr){9-11}
        & & $t0$$\rightarrow$$t1$& $t1$$\rightarrow$$t0$ & TC & $t0$$\rightarrow$$t1$& $t1$$\rightarrow$$t0$ & TC & $t0$$\rightarrow$$t1$& $t1$$\rightarrow$$t0$ & TC \\
        \midrule
        \cellcolor{lightblue}\textit{\textbf{Uni-temporal Training}} & \\
        CSCDNet & \multirow{4}{*}{VL-CMU-CD} &  77.4 & 4.5 & 0.02 & 5.6 & 19.4 & 0.02 & 25.5 & 13.6 & 0.09 & 24.3 \\
        CDResNet & &  74.7 & 2.1 &  0.01 & 2.3 & 4.6 & 0.0 & 25.4 & 13.2 & 0.12 & 20.4 \\
        DR-TANet & & 74.2 & 1.6 & 0.01 & 2.6 & 6.7 & 0.0 & 27.5 & 18.0 & 0.19 & 21.8 \\
        C-3PO & & \textbf{77.6} & 4.6 & 0.02 & 8.0 & 34.3 & 0.02 & 30.4 & 21.0 & 0.16 & 29.3 \\
        \midrule
        CSCDNet & \multirow{5}{*}{TSUNAMI} & 17.0 & 22.1 & 0.40 &  83.6 &  60.3 &  0.46 & 27.8 & 30.5 & 0.29 & 40.2 \\
        CDResNet & & 13.2 & 18.4 & 0.42 & 82.8 &  51.5 &  0.38 & 26.3 & 29.0 & 0.31 & 36.9 \\
        DR-TANet & & 14.8 & 17.3 & 0.53 &  82.0 &  57.6 &  0.44 & 24.4 & 26.8 & 0.35 & 37.2 \\
        C-3PO &  & 27.0 & 27.0 & 1.0 &  \textbf{84.2} &  \textbf{84.2} &  1.0 & 34.3 & 34.3 & 1.0 & 48.5 \\
        \midrule
        CSCDNet & \multirow{5}{*}{ChangeSim} & 20.1 & 3.5 & 0.04 & 2.5 & 6.6 & 0.01 & 43.1 &  18.9 &  0.31 & 15.8 \\
        CDResNet & & 17.5 & 3.1 & 0.02 & 4.6 & 9.9 & 0.01 &  41.3 & 18.2 &  0.18 & 15.8 \\
        DR-TANet & & 20.3 & 4.9 & 0.05 & 5.0 & 8.3 & 0.01 &  40.3 &  18.0 &  0.18 & 16.1 \\
        C-3PO & & 1.6 & 1.6 & 1.0 & 0.3 & 0.3 & 1.0 & 15.8 &  15.8 &  1.0 & 5.9 \\

        \midrule
        \cellcolor{apricot}\textit{\textbf{Bi-temporal Training}} & \\
        CSCDNet & \multirow{4}{*}{VL-CMU-CD} & 64.2 & 64.8 & 0.61 & 8.5 & 7.5 & 0.12 & 24.0 & 23.9 & 0.42 & 32.2 \\
        CDResNet & & 60.6 & 61.7 & 0.58 & 5.0 & 3.8 & 0.18 & 17.7 & 14.1 & 0.29 & 27.2 \\
        DR-TANet &  & 45.6 & 61.9 & 0.40 & 3.5 & 1.2 & 0.02 & 18.0 & 23.0 & 0.26 & 25.5 \\
        C-3PO &  & 58.4 & 63.5 & 0.63 & 2.5 & 1.4 & 0.07 & 21.2 & 23.8 & 0.51 & 28.5 \\
        \midrule
        CSCDNet & \multirow{5}{*}{TSUNAMI}  & 20.9 & 21.4 & 0.56 & 82.4 & 82.8 &  0.82 & 30.0 & 29.4 & 0.46 & 44.5 \\
        CDResNet &  & 17.6 & 17.1 & 0.63 & 81.7 & 81.8 & 0.81 & 28.9 & 29.0 & 0.48 & 42.7 \\
        DR-TANet &  & 18.1 & 17.4 & 0.61 & 81.1 & 82.0 & 0.76 & 28.6 & 28.7 & 0.46 & 42.7 \\
        C-3PO &   & 27.5 & 27.5 & 1.0 & 84.1 & 84.1 & 1.0 & 33.7 & 33.7 & 1.0 & 48.4 \\
        \midrule
        CSCDNet & \multirow{5}{*}{ChangeSim}  & 0.2 & 0.1 & 0.86 & 4.7 & 1.0 & 0.69 & 19.5 & 18.6 &  0.49 & 7.4 \\
        CDResNet &  & 4.0 & 2.0 & 0.75 & 1.1 & 0.5 & 0.53 & 19.8 & 20.7 &  0.48 & 8.0 \\
        DR-TANet & & 7.5 & 3.7 & 0.48 & 2.6 & 1.8 & 0.13 & 27.9 & 21.4 &  0.32 & 10.8 \\
        C-3PO &  & 0.4 & 0.4 & 1.0 & 1.2 & 1.2 & 1.0 & 18.0 & 18.0 &  1.0 & 6.5 \\
        \midrule
        \textbf{GeSCF (Ours)} & Zero-shot & 75.4 & \textbf{75.4}  & \textbf{1.0} & 72.8 & 72.8 & \textbf{1.0} & \textbf{54.8} & \textbf{54.8} & \textbf{1.0} & \textbf{67.7} \\
        \bottomrule
    \end{tabular}
    \caption{\textbf{Quantitative results (F1-score) on standard SCD datasets.} } 
\label{tab:comparison1}
\end{table*}

\begin{table*}[t]
    \footnotesize
    \centering
    \setlength{\tabcolsep}{6.2pt}
    \renewcommand{\arraystretch}{0.5} 
    \begin{tabular}{@{}>{\raggedright}m{3.5cm} @{}>{\centering\arraybackslash}m{2cm} *{10}{>{\centering\arraybackslash}m{0.75cm}}@{}}
            \toprule
            \multirow{3}{*}{\textbf{Method}} & \multirow{3}{*}{\textbf{Training}} & \multicolumn{3}{c}{\makecell{SF-XL (U)}}& \multicolumn{3}{c}{St Lucia (S)} & \multicolumn{3}{c}{\makecell{Nordland (R)}} & \multirow{3}{*}{\textbf{Avg.}} \\
            \cmidrule(lr){3-5} \cmidrule(lr){6-8} \cmidrule(lr){9-11}
               & & $t0$$\rightarrow$$t1$& $t1$$\rightarrow$$t0$ & TC & $t0$$\rightarrow$$t1$& $t1$$\rightarrow$$t0$ & TC & $t0$$\rightarrow$$t1$& $t1$$\rightarrow$$t0$ & TC \\
            \midrule
            \cellcolor{lightblue}\textit{\textbf{Uni-temporal Training}} & \\
            CSCDNet& \multirow{4}{*}{VL-CMU-CD} & 28.6 & 23.1 & 0.07 & 11.7 & 15.3 & 0.27 & 14.3  & 5.8 & 0.33 & 16.5 \\
            CDResNet& & 22.9 & 18.9 & 0.08 & 13.5 & 18.0 & 0.18 & 9.3 & 6.7 & 0.39 & 14.9 \\
            DR-TANet& & 22.5 & 19.6 & 0.05 & 15.5 & 20.8 & 0.19 & 11.8 & 8.2 & 0.33 & 16.4 \\
            C-3PO& & 38.7 & 32.8 & 0.07 & 21.7 & 25.7 & 0.14 & 16.2 & 14.2 & 0.28 & 24.9 \\
            \midrule
            CSCDNet& \multirow{5}{*}{TSUNAMI} & 38.9 & 39.6 & 0.55 & 15.3 & 15.4 & 0.55 & 18.8 & 20.8 & 0.35 & 24.8 \\
            CDResNet&& 36.6 & 37.4 & 0.59 & 13.7 & 14.0 & 0.66 & 16.4 & 19.9 & 0.33 & 23.0 \\
            DR-TANet&& 35.2 & 36.1 & 0.62 & 13.4 & 13.4 & 0.76 & 17.1 & 16.7 & 0.44 & 22.0 \\
            C-3PO   && 50.0 & 50.0 & 1.0 & 28.4 & 28.4 & 1.0 & 23.2 & 23.2 & 1.0 & 33.9 \\
            \midrule
            CSCDNet& \multirow{5}{*}{ChangeSim} & 32.8 & 28.1 & 0.08 & 19.3 & 19.6 & 0.09 & 6.7 & 7.8 & 0.29 & 19.1 \\
            CDResNet&& 26.6 & 23.4 & 0.07 & 15.2 & 14.2 & 0.15 & 7.1 & 6.4 & 0.3 & 15.5 \\
            DR-TANet&& 27.5 & 25.0 & 0.08 & 15.9 & 15.9 & 0.15 & 10.6 & 13.8 & 0.21 & 18.1 \\
            C-3PO   && 4.0 & 4.0 & 1.0 & 3.1 & 3.1 & 1.0 & 0.8 & 0.8 & 1.0 & 2.6 \\
            \midrule

            \cellcolor{apricot}\textit{\textbf{Bi-temporal Training}} & \\
            CSCDNet & \multirow{4}{*}{VL-CMU-CD} & 24.1 & 25.4 & 0.40 & 5.6 & 5.1 & 0.68 & 2.2 & 3.0 & 0.71 & 10.9 \\
            CDResNet && 23.1 & 23.3 & 0.37 & 7.0 & 6.7 & 0.69 & 1.3 & 1.2 & 0.80 & 10.4 \\
            DR-TANet && 17.0 & 17.3 & 0.27 & 9.0 & 4.1 & 0.56 & 2.2 & 2.4 & 0.66 & 8.7 \\
            C-3PO    && 23.3 & 23.2 & 0.5 & 7.9 & 6.9 & 0.77 & 3.6 & 4.6 & 0.79 & 11.6 \\
            \midrule
            CSCDNet & \multirow{5}{*}{TSUNAMI} & 42.8 & 42.9 & 0.67 & 17.1 & 17.1 & 0.64 & 20.4 & 21.0  & 0.57 & 26.9 \\
            CDResNet && 40.0 & 40.1 & 0.71 & 13.9 & 13.8 & 0.76 & 19.8 & 19.5 & 0.58 & 24.5 \\
            DR-TANet && 38.6 & 38.7 & 0.71 & 13.9 & 14.0 & 0.78 & 20.6 & 17.5 & 0.55 & 23.9 \\
            C-3PO    && 50.0 & 50.0 & 1.0 & 28.0 & 28.0 & 1.0 & 24.3 & 24.3 & 1.0 & 34.1 \\
            \midrule
            CSCDNet & \multirow{5}{*}{ChangeSim} & 13.3 & 12.1 & 0.31 & 0.9 & 1.1 & 0.82 & 0.7 & 1.0 & 0.84 & 4.9 \\
            CDResNet && 11.4 & 9.6 & 0.26 & 2.2 & 1.5 & 0.76 & 1.3 & 0.7 & 0.92 & 4.5 \\
            DR-TANet && 12.8 & 12.5 & 0.12 & 8.3 & 6.1 & 0.30 & 7.5 & 6.3 & 0.40 & 8.9 \\
            C-3PO    && 6.0 & 6.0 & 1.0 & 2.1 & 2.1 & 1.0 & 1.0  & 1.0 & 1.0 & 3.0 \\
            \midrule
            \textbf{GeSCF (Ours)} & Zero-shot & \textbf{71.2} & \textbf{71.2} & \textbf{1.0} & \textbf{62.1} & \textbf{62.1} & \textbf{1.0} & \textbf{59.0} & \textbf{59.0} & \textbf{1.0} & \textbf{64.1} \\
            \bottomrule
    \end{tabular}
    \caption{\textbf{Quantitative results (F1-score) on the ChangeVPR dataset.}} 
\label{tab:comparison2}
\end{table*}


\begin{figure}[t]
    \centering
    \includegraphics[width=\linewidth]{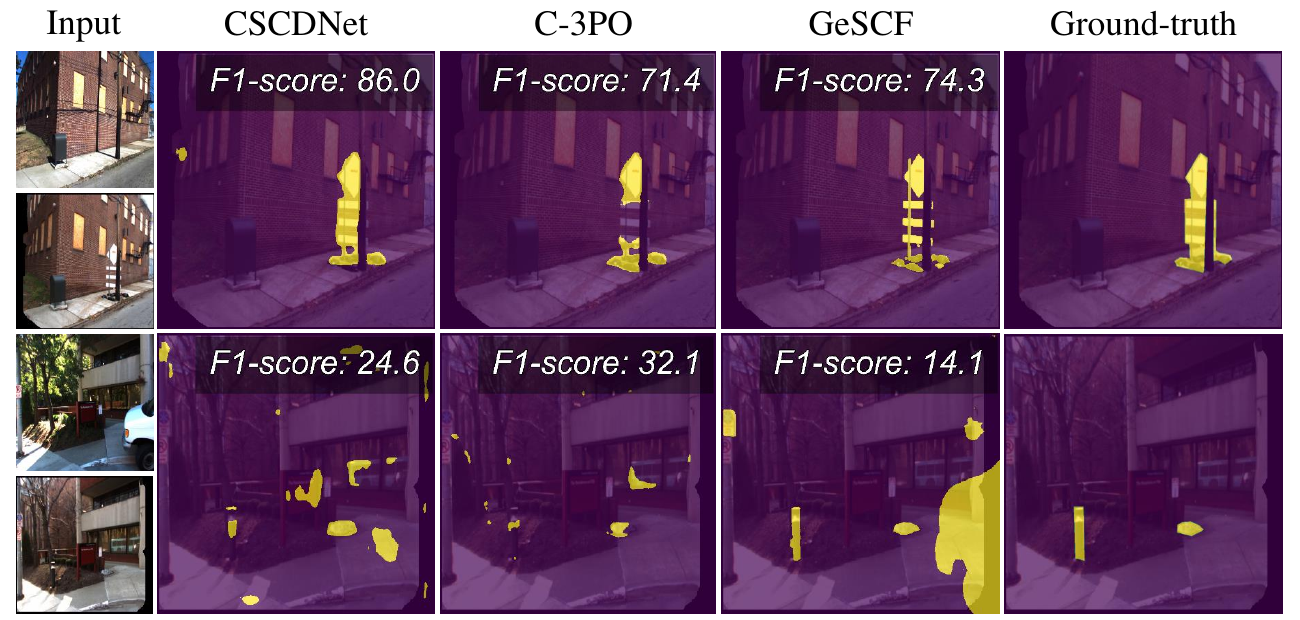}
    \caption{\textbf{Qualitative results on VL-CMU-CD dataset with F1-scores.} Our model generates reasonable change masks and does not display annotation bias.}
    \label{fig:bias}
\end{figure}

\subsection{Comparative Studies}
We compare our method with four state-of-the-art SCD models: CSCDNet \cite{Sakurada2018WeaklySS}, CDResNet \cite{Sakurada2018WeaklySS}, DR-TANet \cite{Chen2021DRTANetDR}, and C-3PO \cite{wang2023reduce}. For C-3PO, we adopt the (I+D) structure for the VL-CMU-CD and the (I+A+D+E) structure for the TSUNAMI and ChangeSim datasets, following the configurations proposed in the paper. To verify the temporal consistency, we additionally train all models with bi-temporal objective \cite{Zheng2021ChangeIE}; the performance of these bi-temporally trained models has not been reported in the literature.

\noindent\textbf{Quantitative Comparison.} \cref{tab:comparison1} displays the quantitative comparison results on the standard SCD datasets. The results attest that GeSCF outperforms the baselines with a large margin in every unseen domain (off-diagonal results) and on-par performance on seen domains (diagonal results), along with complete TC score (TC$=$$1.0$) for all settings---demonstrating its superior generalizability over diverse environments and temporal conditions. The current best performance on the VL-CMU-CD dataset drops sharply from $77.6\%$ to $4.6\%$ when the temporal order is reversed and to $8.0\%$ when deployed to the unseen TSUNAMI dataset. Remarkably, our GeSCF even outperforms the supervised baselines for the ChangeSim dataset by an F1-score of 54.8\% in a zero-shot manner. Overall, the average performance of our method on standard SCD datasets surpasses that of the second-best by a substantial margin ($+19.2\%$). Furthermore, the majority of baselines do not display a complete TC score which is crucial for the system's reliability. The temporal consistency of C-3PO does not hold in VL-CMU-CD settings as shown in its low TC scores. We find that simple bi-temporal training does not guarantee complete temporal consistency and unseen domain performance, even worsening the uni-temporal performances---indicating different data characteristics from \cite{Zheng2021ChangeIE} and the significance of our commutative architecture. \cref{tab:comparison2} presents the quantitative comparison results on the proposed ChangeVPR dataset. The results consistently show that GeSCF outperforms the baselines across all unseen domains with a $+30.0\%$ margin (\textit{nearly doubling} the prior art performance), extending beyond the conventional urban-/synthetic-only SCD scope. Notably, GeSCF demonstrates exceptional performance in Nordland \cite{Snderhauf2013AreWT} split, a challenging dataset with severe seasonal variation (summer-winter).

\noindent\textbf{Qualitative Comparison.} As illustrated in \cref{fig:title}, GeSCF adeptly adapts to various unseen images, accurately delineating changed objects---significantly outperforming the current state-of-the-art model. Moreover, as a zero-shot framework, GeSCF does not learn annotation biases in a dataset (see \cref{fig:bias}). For instance, GeSCF generates a more accurate and interpretable mask for the object and identifies unannotated changes---displaying resilience to erroneous GTs. These results suggest that GeSCF's outputs are more reasonable than baselines, even though its quantitative results are lower for some samples in the in-domain settings. 

\subsection{Ablation Studies}
To understand the effectiveness of our contributions, we conducted comprehensive ablation studies across three standard SCD datasets and ChangeVPR (see \cref{tab:adap}). We utilized baselines trained on TSUNAMI \cite{Sakurada2015ChangeDF}, which demonstrated superior average performance over others.

\noindent\textbf{Fine-tuning Strategy.} We fine-tuned various SCD-based adapter networks on the frozen SAM image encoder, involving correlation layers \cite{Dosovitskiy2015FlowNetLO}, CHVA \cite{Chen2021DRTANetDR}, and the feature merging module \cite{wang2023reduce}. Our experiments reveal that fine-tuning the adaptor model on relatively small SCD datasets significantly undermines its zero-shot capabilities, resulting in a substantial performance drop compared to our proposed GeSCF. Notably, our initial pseudo-masks alone achieve an F1-score that surpasses the fine-tuned variants.


\noindent\textbf{Effectiveness of the Proposed Modules.} We validated the effectiveness of each proposed module, including the Initial Pseudo-mask Generation with adaptive threshold function, Geometric-Semantic Mask Matching with geometric intersection matching (GIM), and semantic similarity matching (SSM). The findings suggest that each introduced module positively influences GeSCF's overall performance, with the initial pseudo-mask generation (adaptive) and GIM demonstrating significant enhancement.

\begin{table}[t]
    \footnotesize
    \centering
    \setlength{\tabcolsep}{6.2pt}
    \renewcommand{\arraystretch}{0.9} 
      \begin{tabular}{@{}>{\raggedright}m{4cm} *{2}{>{\centering\arraybackslash}m{1.1cm}}@{}}
            \toprule
            \multirow{3}{*}{\textbf{Method}}  & \multicolumn{2}{c}{\textbf{Metric}} \\
            \cmidrule(lr){2-3}
             & F1-score & mIoU \\
            \midrule
            \cellcolor{lightblue} \textit{\textbf{Uni-temporal Training}} & & \\
            SAM \cite{Kirillov2023SegmentA} + Corr. layers \cite{Dosovitskiy2015FlowNetLO} & 28.7 & 19.7 \\
            SAM \cite{Kirillov2023SegmentA} + CHVA \cite{Chen2021DRTANetDR} & 30.4 & 21.1 \\
            SAM \cite{Kirillov2023SegmentA} + MTF \cite{wang2023reduce} + MSF \cite{wang2023reduce} & 40.1 & 30.2 \\
            \midrule
            \cellcolor{apricot}\textit{\textbf{Bi-temporal Training}} & & \\
            SAM \cite{Kirillov2023SegmentA} + Corr. layers \cite{Dosovitskiy2015FlowNetLO} & 30.0 & 21.1 \\
            SAM \cite{Kirillov2023SegmentA} + CHVA \cite{Chen2021DRTANetDR} & 31.0 & 22.0 \\
            SAM \cite{Kirillov2023SegmentA} + MTF \cite{wang2023reduce} + MSF \cite{wang2023reduce} & 40.3 & 30.3 \\
            \midrule
            \cellcolor{myLightGray}\textit{\textbf{GeSCF (Ours)}} & & \\
            Pseudo. only (fixed) & 43.0 & 26.3 \\
            Pseudo. only (adaptive) & 53.9 & 34.9 \\
            Pseudo. + GIM & 65.0 & 47.6 \\
            Pseudo. + GIM + SSM (\textbf{ALL}) & \textbf{65.9} & \textbf{48.8}  \\
            \bottomrule
      \end{tabular}
\caption{\textbf{Ablation study results.}}
\label{tab:adap}
\end{table}

\begin{table}[t]
    \footnotesize
    \centering
    \setlength{\tabcolsep}{6.2pt}
    \renewcommand{\arraystretch}{0.9} 
      \begin{tabular}{@{}>{\raggedright}m{2.5cm} *{3}{>{\centering\arraybackslash}m{1cm}}@{}}
            \toprule
            \multirow{3}{*}{\textbf{Method}}  & \multicolumn{3}{c}{\textbf{Metric}} \\
            \cmidrule(lr){2-4}
             & F1-score & Precision & Recall \\
            \midrule
            ISFA \cite{Wu2014SlowFA}     & 32.9 & 29.8 & 36.8 \\
            DSFA \cite{Du2018UnsupervisedDS}     & 33.0 & 24.2 & \textbf{51.9} \\
            DCAE \cite{Bergamasco2022UnsupervisedCD}    & 33.4 & 35.7 & 31.4 \\
            OBCD \cite{Xiao2016ChangeDO}   & 34.3 & 29.6 & 40.7 \\
            KPCA-MNet \cite{Wu2019UnsupervisedCD}    & 36.7 & 29.5 & 48.5 \\
            DCVA \cite{Saha2019UnsupervisedDC}    & 36.8 & 29.6 & 48.7 \\
            \midrule
            \textbf{GeSCF (Ours)}         & \textbf{48.0} & \textbf{52.2} & 44.4 \\
            \bottomrule
      \end{tabular}
\caption{\textbf{Quantitative comparison with unsupervised remote sensing CD methods on the SECOND (\texttt{test}) benchmark.}}
\label{tab:remote}
\end{table}

\subsection{Beyond Scene Change Detection}
Although our research primarily focuses on natural scene CDs and each CD field tends to focus on its own \cite{Lee2024SemiSupervisedSC}, we discovered that GeSCF can also be applied to zero-shot remote sensing CD. Following the standard evaluation metrics of the remote sensing domain, we evaluated our model against several unsupervised methods \cite{Wu2014SlowFA,Du2018UnsupervisedDS,Bergamasco2022UnsupervisedCD,Xiao2016ChangeDO,Wu2019UnsupervisedCD,Saha2019UnsupervisedDC} on the SECOND \cite{Yang2022AsymmetricSN} benchmark (see \cref{tab:remote}). Remarkably, our GeSCF outperforms other baselines with 48.0\% in F1-score and a high precision of 52.2\%---underscoring its zero-shot potential for different CD domains.


\section{Conclusion}

In this study, we attempted to address the generalization issue and temporal bias of scene change detection using the Segment Anything Model for the first time to the best of our knowledge; we define GeSCD along with the novel metrics, evaluation dataset (ChangeVPR), and an evaluation protocol to effectively assess the generalizability and temporal consistency of SCD models. Furthermore, we proposed GeSCF, a generalizable zero-shot approach for the SCD task using the localized semantics of the SAM that does not require costly SCD labeling. Our extensive experiments demonstrated that the proposed GeSCF significantly outperforms existing SCD models on standard SCD datasets and overwhelms them on the ChangeVPR dataset while achieving complete temporal consistency. We expect our methods can serve as a solid step towards robust and generalizable Scene Change Detection research.

\fontsize{8}{10}\selectfont{%
\noindent\textbf{Acknowledgement.}
This research was partly supported by the Institute of Information \& communications Technology Planning \& Evaluation (IITP) grants funded by the Korea government (MSIT) (No.RS-2022-II220926, Development of Self-directed Visual Intelligence Technology Based on Problem Hypothesis and Self-supervised Methods; No.RS-2022-II220907, Development of AI Bots Collaboration Platform and Self-organizing AI; No.2019-0-01842, Artificial Intelligence Graduate School Program (GIST)), National Research Foundation of Korea (NRF) grant funded by the Korea government (MSIT) (No.NRF-2022R1C1C1009989), and 'Project for Science and Technology Opens the Future of the Region' program through INNOPOLIS FOUNDATION founded by Ministry of Science and ICT (No.2022-DD-UP-0312-02-101).

{
    \small
    \bibliographystyle{ieeenat_fullname}
    \bibliography{main}
}


\end{document}


\clearpage
\setcounter{page}{1}
\maketitlesupplementary

\section{ChangeVPR Dataset}
\begin{figure}[t!]
    \centering
    \includegraphics[width=\linewidth]{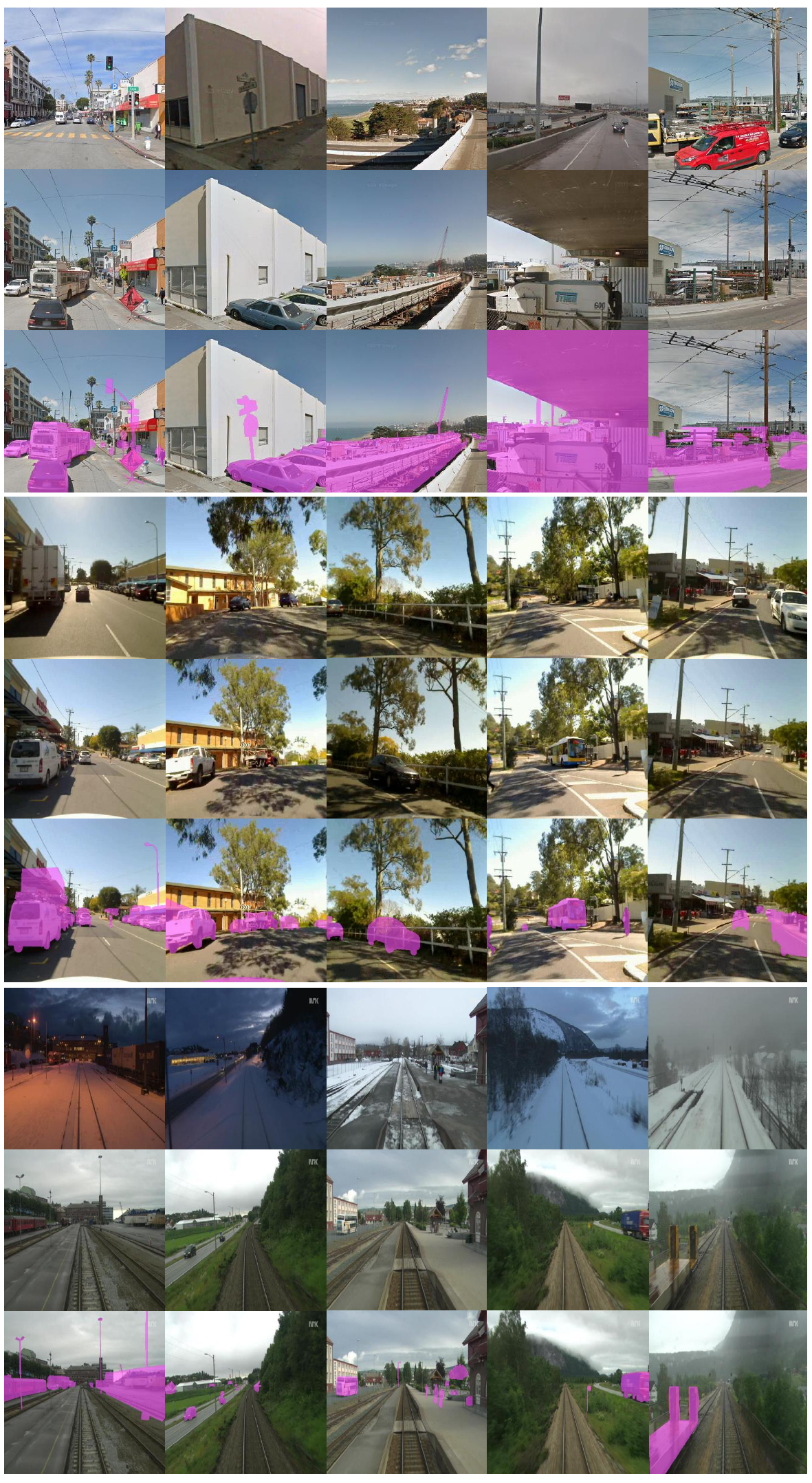}
        \caption{\textbf{Examples of ChangeVPR dataset.} (Top) SF-XL. (Middle) St Lucia. (Bottom) Nordland. Each row of the splits represents a query, reference, and ground-truth mask.}
    \label{examples}
\end{figure}

The meticulously annotated ChangeVPR dataset is designed to evaluate the robustness and generalizability of scene change detection models. It covers a diverse range of environments with challenging scenarios---significantly expanding the traditional urban or synthetic-only SCD scope. We source the images from three widely used visual place recognition datasets with different environmental characteristics: SF-XL \cite{Berton2022RethinkingVG} (urban), St Lucia \cite{Milford2008MappingAS} (suburban), and Nordland \cite{Snderhauf2013AreWT} (rural) dataset (see \cref{examples}). The detailed descriptions of each split are as follows:

\begin{itemize}
    \item \textbf{SF-XL} is a vast dataset covering San Francisco, with over 41M images collected from Google Street View. The temporal distribution of the dataset ranges from 2007 to 2020. We quantize the whole dataset using classification strategy \cite{Berton2022RethinkingVG} and carefully select two bi-temporal images from quantized cells for query and reference. Following \cite{Berton2022RethinkingVG}, we consider two bi-temporal images to be of the same place (same class) if they are located in the same geographical cell (a cell with a side of 10 meters) and their heading difference is less than 30 degrees.
    
    \item \textbf{St Lucia} features video recordings from a car-mounted camera, capturing multiple drives through the St Lucia suburb of Brisbane. Following \cite{Berton2022DeepVG}, among nine available drives, we use the first drive as a query and the last as reference sets. Since there are no heading labels for the image, we carefully select image pairs after sampling two bi-temporal images with a UTM distance of less than 10 meters.
    
    \item \textbf{Nordland} captures a train journey through the Norwegian countryside, capturing the same route across four seasons with frames extracted at 1 FPS. Following previous VPR works \cite{Hausler2019MultiProcessFV,Hausler2021PatchNetVLADMF}, which use the winter traverse as queries and summer as a database, we adopt winter traverse as a query and summer as a reference. Since no heading labels exist, we follow the same process in the St Lucia split.
    
\end{itemize}

\noindent We manually labeled the ground-truth for scene changes, providing it as a binary image, matching the size of the input image pairs of resolution 512$\times$512. In this binary image, each pixel value indicates whether a change occurred at the corresponding point between the bi-temporal images. Following the convention \cite{Sakurada2015ChangeDF,Alcantarilla2016StreetviewCD}, we define scene changes as both 2D surface alterations (e.g., changes to advertising boards) and 3D structural modifications (e.g., the appearance or disappearance of buildings, vehicles, trash bins, and pedestrians). Finally, the ChangeVPR dataset contains the binary change masks $C_{t0}$, $C_{t1}$ and the intersection change mask $C_{t0\leftrightarrow t1}$, respectively.

\begin{table*}[t]
    \footnotesize
    \centering
    \setlength{\tabcolsep}{6.2pt}
    \renewcommand{\arraystretch}{0.5} 
    \begin{tabular}{@{}>{\raggedright}m{2cm} @{}>{\centering\arraybackslash}m{2cm} *{10}{>{\centering\arraybackslash}m{0.6cm}}@{}}
        \toprule
        \multirow{3}{*}{\textbf{Method}} & \multirow{3}{*}{\textbf{Training}} & \multicolumn{3}{c}{VL-CMU-CD} & \multicolumn{3}{c}{TSUNAMI} & \multicolumn{3}{c}{ChangeSim} & \multirow{3}{*}{\textbf{Avg.}}  \\
        \cmidrule(lr){3-5} \cmidrule(lr){6-8} \cmidrule(lr){9-11}
        & & $t0$$\rightarrow$$t1$& $t1$$\rightarrow$$t0$ & TC & $t0$$\rightarrow$$t1$& $t1$$\rightarrow$$t0$ & TC & $t0$$\rightarrow$$t1$& $t1$$\rightarrow$$t0$ & TC \\
        \midrule
        \multirow{2.8}{*}{SimSaC} & VL-CMU-CD$^{\dagger}$ & \textbf{75.5} & 42.3 & 0.39 & 32.0 & 41.9 & 0.27 & 53.5 &  32.2 &  0.24 & 46.2 \\
        & ChangeSim$^{\dagger}$ & 74.1 & 17.7 & 0.12 & 17.7 & 20.6 & 0.13 & \textbf{62.3} & 37.3 & 0.29 & 38.2 \\
        \midrule
        \rowcolor{LightCyan}
        \textbf{GeSCF (Ours)} & Zero-shot & 75.4 & \textbf{75.4}  & \textbf{1.0} & \textbf{72.8} & \textbf{72.8} & \textbf{1.0} & 54.8 & \textbf{54.8} & \textbf{1.0} & \textbf{67.7} \\
        \bottomrule
    \end{tabular}
    \caption{\textbf{Quantitative results of SimSaC on standard SCD datasets.} $^{\dagger}$ indicates that an additional Synthetic \cite{Park2022DualTL} dataset is used for training.}
\label{sup:comparison_simsac_scd}
\end{table*}

\begin{table*}[t]
    \footnotesize
    \centering
    \setlength{\tabcolsep}{6.2pt}
    \renewcommand{\arraystretch}{0.5} 
    \begin{tabular}{@{}>{\raggedright}m{2cm} @{}>{\centering\arraybackslash}m{2cm} *{10}{>{\centering\arraybackslash}m{0.6cm}}@{}}
        \toprule
        \multirow{3}{*}{\textbf{Method}} & \multirow{3}{*}{\textbf{Training}} & \multicolumn{3}{c}{SF-XL (U)} & \multicolumn{3}{c}{St Lucia (S)} & \multicolumn{3}{c}{Nordland (R)} & \multirow{3}{*}{\textbf{Avg.}}  \\
        \cmidrule(lr){3-5} \cmidrule(lr){6-8} \cmidrule(lr){9-11}
        & & $t0$$\rightarrow$$t1$& $t1$$\rightarrow$$t0$ & TC & $t0$$\rightarrow$$t1$& $t1$$\rightarrow$$t0$ & TC & $t0$$\rightarrow$$t1$& $t1$$\rightarrow$$t0$ & TC \\
        \midrule
        \multirow{2.8}{*}{SimSaC} & VL-CMU-CD$^{\dagger}$ & 56.3 & 55.0 & 0.31 & 51.2 & 51.8 & 0.32 & 28.1 & 25.5 & 0.25 & 44.7 \\
        & ChangeSim$^{\dagger}$ & 44.3 & 40.5 & 0.12 & 34.3 & 36.7 & 0.04 & 19.4 & 17.5 & 0.05 & 32.1 \\
        \midrule
        \rowcolor{LightCyan}
        \textbf{GeSCF (Ours)} & Zero-shot & \textbf{71.2} & \textbf{71.2} & \textbf{1.0} & \textbf{62.1} & \textbf{62.1} & \textbf{1.0} & \textbf{59.0} & \textbf{59.0} & \textbf{1.0} & \textbf{64.1} \\
        \bottomrule
    \end{tabular}
    \caption{\textbf{Quantitative results of SimSaC on ChangeVPR dataset.} $^{\dagger}$ indicates that an additional Synthetic \cite{Park2022DualTL} dataset is used for training.}
\label{comparison_simsac_vpr}
\end{table*}

\begin{table*}[t!]
    \footnotesize
    \centering
    \setlength{\tabcolsep}{6.2pt}
    \renewcommand{\arraystretch}{0.5} 
    \begin{tabular}{@{}>{\raggedright}m{2cm} @{}>{\centering\arraybackslash}m{2cm} *{10}{>{\centering\arraybackslash}m{0.6cm}}@{}}
        \toprule
        \multirow{3}{*}{\textbf{Method}} & \multirow{3}{*}{\textbf{Training}} & \multicolumn{3}{c}{VL-CMU-CD} & \multicolumn{3}{c}{TSUNAMI} & \multicolumn{3}{c}{ChangeSim} & \multirow{3}{*}{\textbf{Avg.}}  \\
        \cmidrule(lr){3-5} \cmidrule(lr){6-8} \cmidrule(lr){9-11}
        & & $t0$$\rightarrow$$t1$& $t1$$\rightarrow$$t0$ & TC & $t0$$\rightarrow$$t1$& $t1$$\rightarrow$$t0$ & TC & $t0$$\rightarrow$$t1$& $t1$$\rightarrow$$t0$ & TC \\
        \midrule
        C-3PO &  \multirow{2.8}{*}{VL-CMU-CD} & \textbf{77.4} & 4.5 & 0.02 & 5.6 & 19.4 & 0.02 & 25.5 & 13.6 & 0.09 & 24.3 \\
        C-3PO$^{*}$ && 59.0 & 59.0 & 1.0 & 5.1 & 5.1 & 1.0 & 17.2 & 17.2 & 1.0 & 27.1 \\
        \midrule
        \rowcolor{LightCyan}
        \textbf{GeSCF (Ours)} & Zero-shot & 75.4 & \textbf{75.4}  & \textbf{1.0} & \textbf{72.8} & \textbf{72.8} & \textbf{1.0} & \textbf{54.8} & \textbf{54.8} & \textbf{1.0} & \textbf{67.7} \\
        \bottomrule
    \end{tabular}
    \caption{\textbf{Quantitative results of C-3PO variants on VL-CMU-CD dataset.} $^{*}$ indicates the (I+A+D+E) structure; otherwise, the (I+D) structure for C-3PO.} 
\label{sup:comparison_c3po}
\end{table*}
\

\section{Implementation Details}
\subsection{Hyperparameters of GeSCF}
To ensure general applicability and consistent performance on real-world applications, \textit{we uniformly set the SAM parameters for all configurations}, including the standard SCD datasets and our ChangeVPR dataset. Specifically, we employ SAM ViT-H with a point per side of $32$, an NMS threshold of 0.7, a predicted IoU threshold of 0.7, and a stability score threshold of 0.7. For the adaptive threshold function in GeSCF, we adopt $b_{\gamma}$$=$$0.05$ and $s_{\gamma}$$=$$0.1$ for right-skewed distributions ($\gamma$$>$$0.2$), and $b_{\gamma}$$=$$0.7$ and $s_{\gamma}$$=$$1.0$ for left-skewed distributions ($\gamma$$<$$-0.2$). For moderate distributions ($-0.2$$\leq$$\gamma$$\leq$$0.2$), we employ the z-score method with a z-value of $-0.52$. Further, we intercept key facets from the 17th layer of the SAM ViT-H encoder. In the Geometric-Semantic Mask Matching module, we set $\alpha_t$ to $0.65$ and change confidence score to 0.88. We conduct a linear search on a small validation set sampled from the VL-CMU-CD training set to find the optimal hyperparameters. Importantly, after setting these hyperparameters, GeSCF requires no further tuning for specific datasets—--the same hyperparameters are used consistently across all settings.

\subsection{Adaptor Networks}
In the absence of SAM-based baselines, we have meticulously constructed various adaptor networks based on relevant literature for ablation. As change detection necessitates processing bi-temporal features, adaptor \cite{Xu2023ParameterEfficientFM} networks are commonly used when leveraging foundation models---which usually take single-image inputs---in the CD domain \cite{Ding2023AdaptingSA,Mei2024SCDSAMAS}. To demonstrate our effectiveness of GeSCF in SAM's feature utilization for generalizability through ablation studies, we configured several adaptor networks on the frozen SAM ViT-H image encoder with three prominent feature processing modules in SCD: Correlation layers \cite{Dosovitskiy2015FlowNetLO}, CHVA \cite{Chen2021DRTANetDR}, and feature merging modules \cite{wang2023reduce}. Specifically, following \cite{Ding2023AdaptingSA}, we intercept four image embeddings at different ViT blocks to obtain local and global features from the SAM ViT image encoder \cite{Ghiasi2022WhatDV}. We evenly select features after the global attention layers of the SAM image encoder \cite{Ke2023SegmentAI}; for the ViT-H model, these are the outputs of the 7th, 15th, 23rd, and 31st blocks output for the 32 blocks
in total. The intercepted features are then interpolated to form a multi-scale feature pyramid, which is input to the feature processing modules. The processed image features are subsequently fed to a decoder network, following the configurations in \cite{Sakurada2018WeaklySS, Chen2021DRTANetDR, wang2023reduce}. As shown in the manuscript, leveraging SAM with learnable adapters significantly degrades generalizability, since the current SCD datasets are not representative enough to cover diverse real-world changes.

\begin{figure*}[t]
    \centering
    \includegraphics[width=\linewidth]{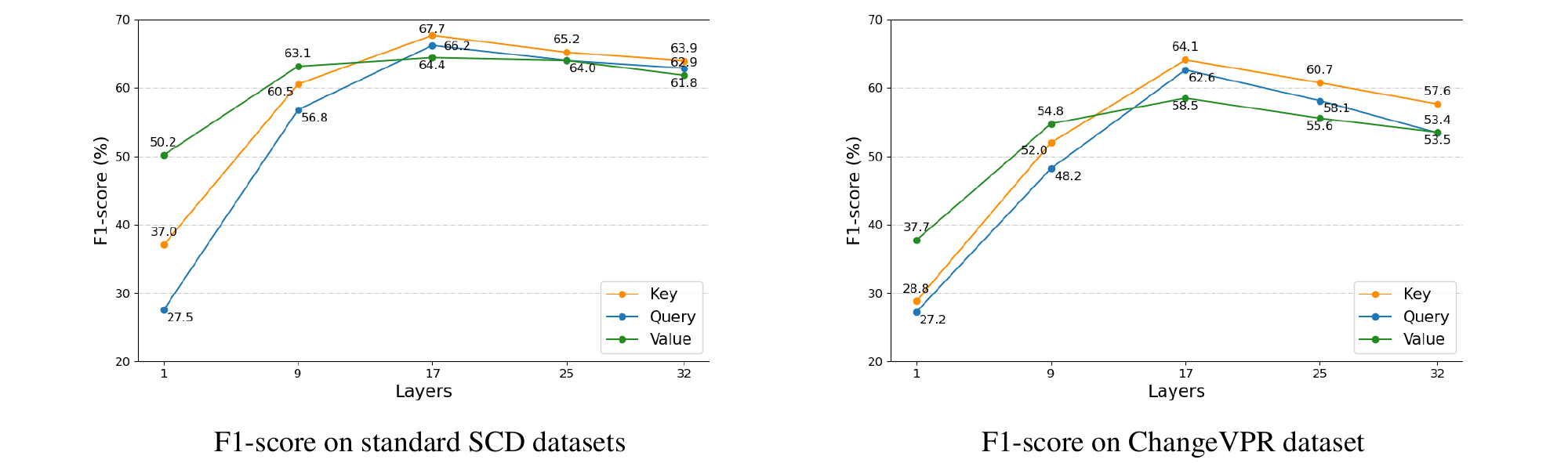}
    \caption{\textbf{Comparative analysis of GeSCF performance using various facets and layers across standard SCD datasets (VL-CMU-CD \cite{Alcantarilla2016StreetviewCD}, TSUNAMI \cite{Sakurada2015ChangeDF}, and ChangeSim \cite{Park2021ChangeSimTE}) and the ChangeVPR dataset.} The key facet from the intermediate layer achieves the best performance among other choices, highlighting that our facet and layer selection is quantitatively a reasonable design choice.}
    \label{sup:facet_quan}
\end{figure*}

\begin{table}[t!]
    \footnotesize
    \centering
    \setlength{\tabcolsep}{6.2pt}
    \renewcommand{\arraystretch}{0.5} 
    \begin{tabular}{@{}>{\raggedright}m{2cm} *{4}{>{\centering\arraybackslash}m{1cm}}@{}}
        \toprule
        \multirow{3}{*}{\textbf{Backbone}} & \multicolumn{2}{c}{SCD Datasets} & \multicolumn{2}{c}{ChangeVPR} \\
        \cmidrule(lr){2-3} \cmidrule(lr){4-5}
        & F1-score & mIoU & F1-score & mIoU \\
        \midrule
        DINOv2 (ViT-B) & 60.8  & 43.6 & 50.7 & 32.8 \\
        SAM (ViT-B) & 61.5 & 42.9 & 53.0 & 34.4  \\        \midrule
        DINOv2 (ViT-L) & 62.3 & 45.3 & 50.9 & 35.9  \\
        SAM (ViT-L) & 64.8 & 47.6 & 59.4 & 40.4  \\
        \midrule
        \rowcolor{LightCyan}
        SAM (ViT-H) & \textbf{67.7} & \textbf{51.3} & \textbf{64.1} & \textbf{46.3} \\
        \bottomrule
    \end{tabular}
    \caption{\textbf{Performance comparison of GeSCF using different ViT backbones.} We maintain the default GeSCF configuration, modifying only the backbone during the initial pseudo-mask generation and semantic similarity matching processes.} 
\label{sup:backbone}
\end{table}

\begin{figure*}[t]
    \centering
    \vspace{0.5cm}
    \includegraphics[width=\linewidth]{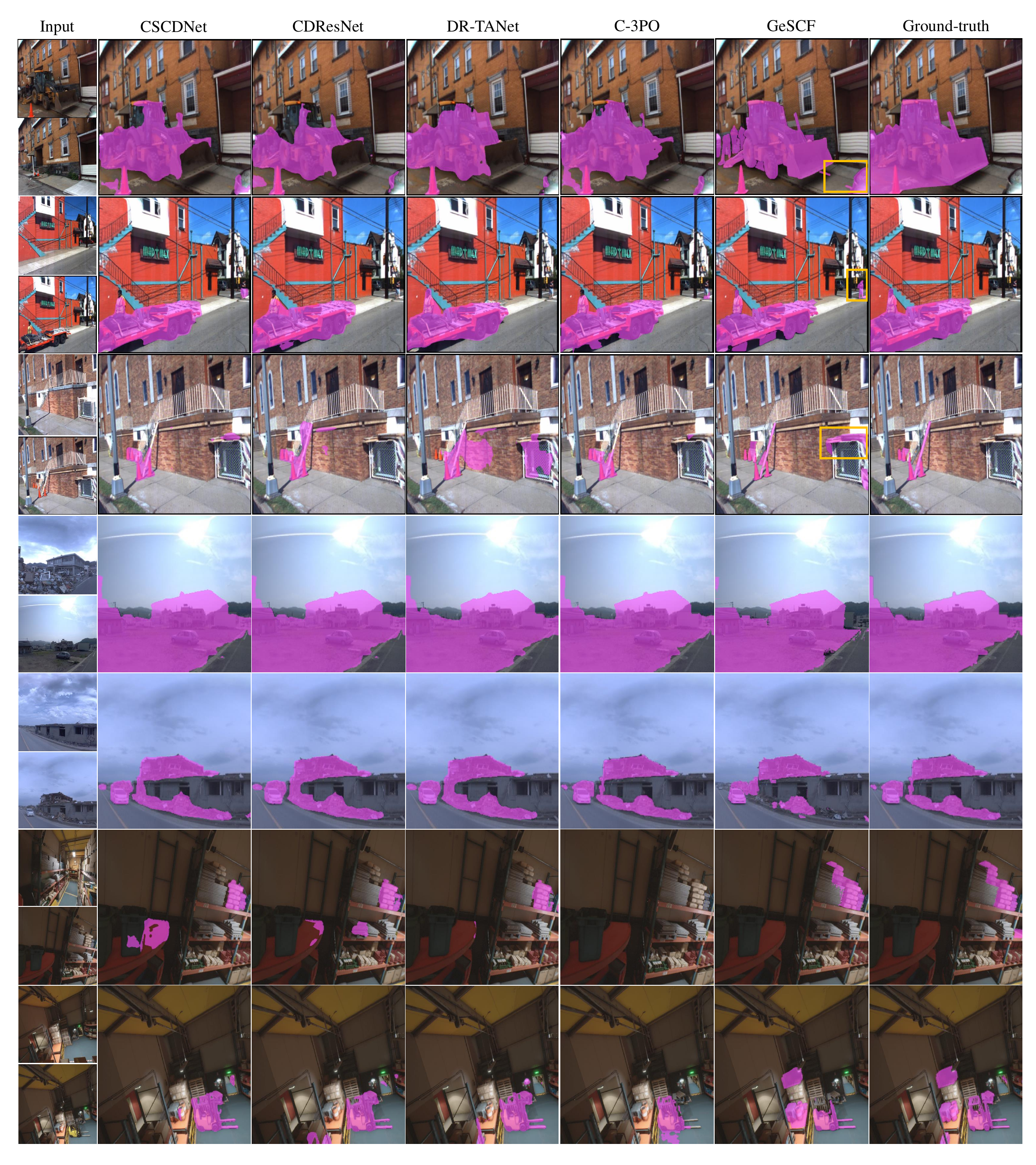}
    \caption{\textbf{More qualitative examples on the seen domain (standard SCD datasets).} Although GeSCF does not learn dataset biases, it can accurately segment meaningful changes with comparable performance or even better than other in-domain baselines. Moreover, our GeSCF can effectively segment unannotated semantic changes highlighted with yellow bounding boxes.}
    \label{sup:more_qual2}
    \vspace{0.5cm} 
\end{figure*}
\begin{figure*}[t]
    \centering
    \vspace{0.5cm}
    \includegraphics[width=\linewidth]{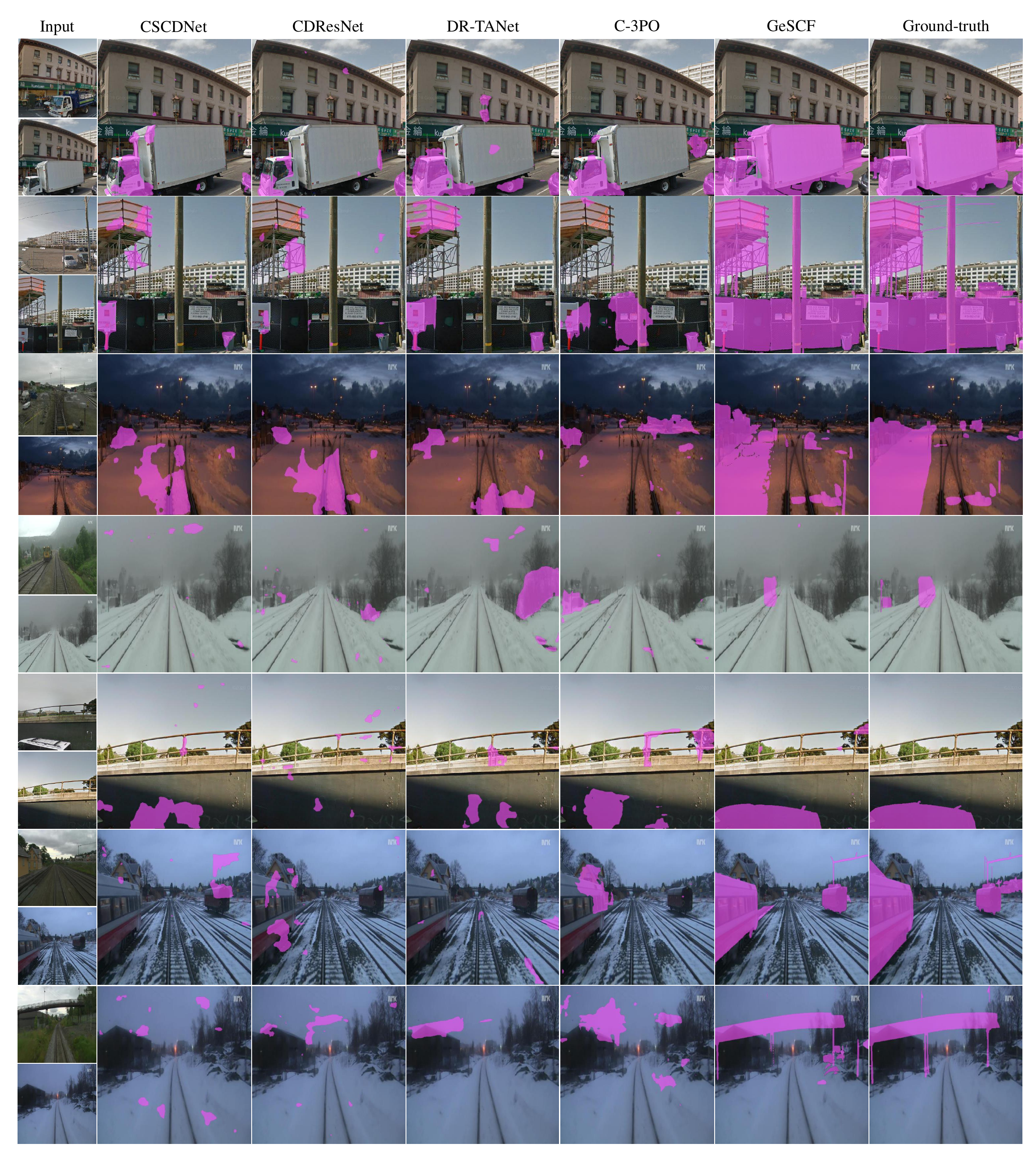}
    \caption{\textbf{More qualitative examples on the unseen domain (ChangeVPR).} Our GeSCF produces accurate and sharp masks with exceptional generalizability, capturing changes across diverse and challenging scenarios more effectively than other existing baselines.}
    \label{sup:more_qual}
    \vspace{0.5cm}
\end{figure*}

\section{More Experiments}

\subsection{More Quantitative Comparisons}
\textbf{Comparison with SimSaC.} In \cref{sup:comparison_simsac_scd,comparison_simsac_vpr}, we provide additional comparison results between our GeSCF and SimSaC \cite{Park2022DualTL}. We were unable to perform full assessments due to the incomplete release of the necessary code and the Synthetic dataset \cite{Park2022DualTL} required for model training. Instead, we leveraged the pre-trained models available on VL-CMU-CD \cite{Alcantarilla2016StreetviewCD} and ChangeSim \cite{Park2021ChangeSimTE}, along with the Synthetic dataset \cite{Park2022DualTL} accessible from the official website. The results align with those presented in the main paper, demonstrating that our GeSCF shows superior robustness on unseen domains and achieves comparable performance on seen domains, all while ensuring complete temporal consistency. Overall, our GeSCF surpasses SimSaC with an exceptional margin, achieving an average improvement of 21.5\% on standard SCD datasets and 19.4\% on the ChangeVPR dataset.


\noindent\textbf{Comparison with C-3PO Variant.}
C-3PO \cite{wang2023reduce} introduces a series of model variants, each incorporating an inductive bias tailored to specific training domains. For example, the (I+D) structure is utilized for the VL-CMU-CD dataset, whereas the (I+A+D+E) structure is employed for the TSUNAMI and ChangeSim datasets. However, applying models with the appropriate prior knowledge in real-world scenarios remains challenging, as it extends beyond the well-explored research datasets. The discrepancy between the model’s built-in assumptions and the application domain can result in suboptimal performance (see \cref{sup:comparison_c3po}). Notably, the performance of C-3PO (I+A+D+E) on the VL-CMU-CD dataset is significantly hindered due to the incorrect assumption, even though it maintains complete temporal consistency. In contrast, our GeSCF provides a robust and unified framework that consistently performs well, maintaining temporal consistency across diverse application domains.


\subsection{More Qualitative Evaluations} 
We present extensive results that demonstrate the effectiveness of our approach across various change scenarios in both seen and unseen domains (see \cref{sup:more_qual2,sup:more_qual}). Unlike conventional SCD methods, which are confined to detecting changes within specific training datasets, our GeSCF captures changes across diverse environments without requiring SCD supervision. Moreover, GeSCF accurately detects meaningful changes beyond dataset biases, achieving performance comparable to or surpassing other supervised in-domain baselines. This capability addresses the longstanding issue of dataset dependency in the SCD field---establishing a robust foundation for a truly applicable and versatile \textit{anything} SCD.

\subsection{Exploring Design Choices in GeSCF}
Our GeSCF builds upon the Segment Anything Model \cite{Kirillov2023SegmentA} (SAM) leveraging the intermediate key facets of SAM ViT image encoder during the initial pseudo-mask generation process, and utilizes the final mask embeddings for the Semantic Similarity Matching (SSM). To validate these design choices, we conduct comprehensive ablation studies.

\noindent\textbf{Key Facets from the Intermediate Layer.} As explained in the main paper, we chose to use key facets of the intermediate layer for the multi-head feature correlation and generate initial pseudo-masks. As we will see here, this choice provides the best change detection performance among other alternatives (see \cref{sup:facet_quan}). While all facets perform best in the intermediate layer, the key facet outperforms the query and value facets in the F1-score. Specifically, on standard SCD datasets, the key facet attains an F1-score of 67.7\%, which is 1.5\% higher than the query facet (66.2\%) and 3.3\% higher than the value facet (64.4\%). Similarly, on the ChangeVPR dataset, utilizing the key facet results in an F1-score of 64.1\%, outperforming the query facet by 1.5\% (62.6\%) and the value facet by 5.6\% (58.5\%). These results emphasize that our selection strategy is quantitatively a well-justified choice.

\begin{figure}[t!]
    \centering
    \includegraphics[width=\linewidth]{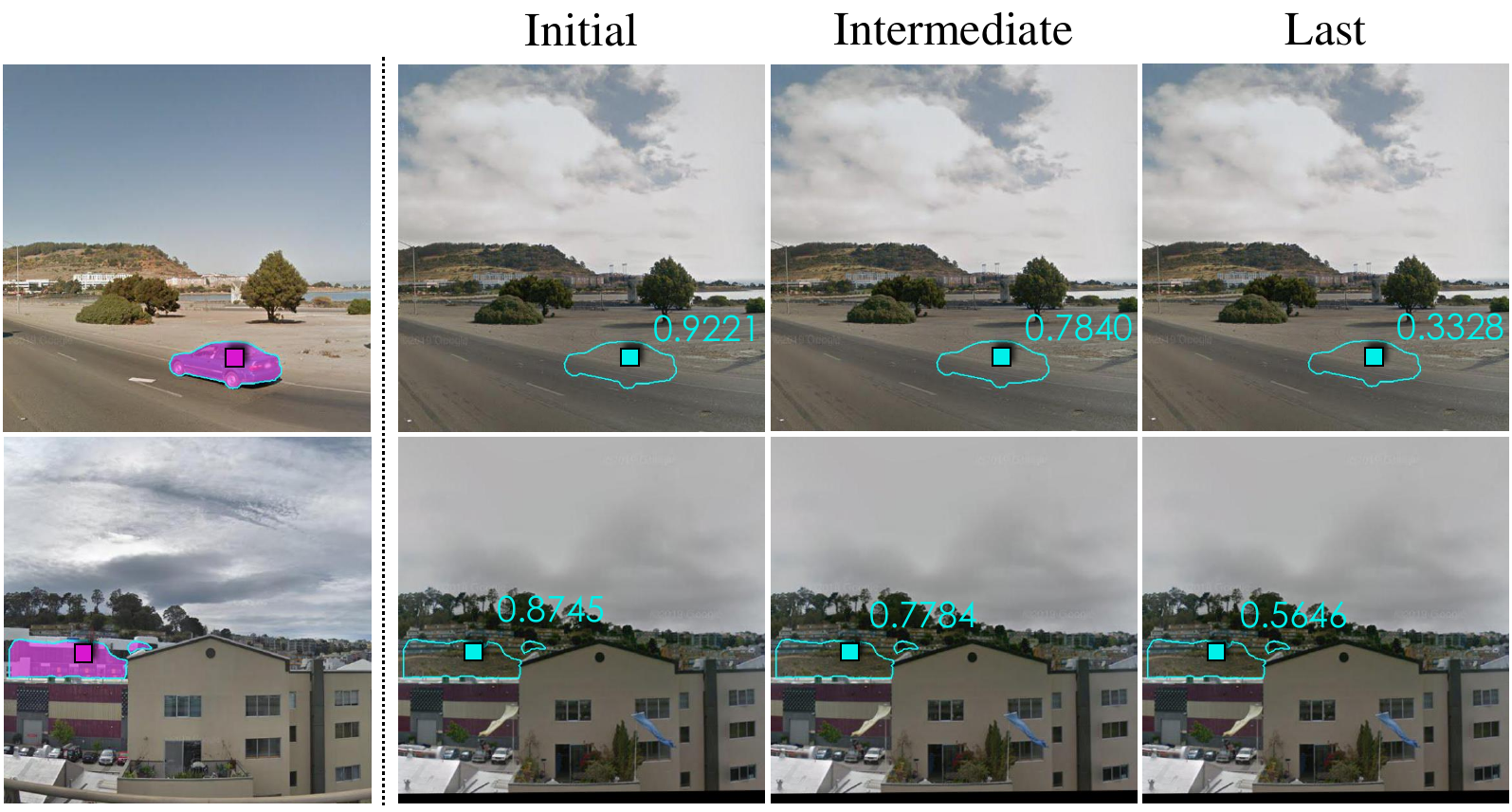}
    \caption{\textbf{Cosine similarity between bi-temporal mask embeddings depending on the layer.} The object-level semantic difference is more pronounced in the last layer compared to the initial and intermediate layers.}
    \label{sup:mask}
\end{figure}

\begin{table}[t!]
    \footnotesize
    \centering
    \setlength{\tabcolsep}{6.2pt}
    \renewcommand{\arraystretch}{0.5} 
    \begin{tabular}{@{}>{\raggedright}m{1.5cm} @{}>{\centering\arraybackslash}m{0.8cm} *{4}{>{\centering\arraybackslash}m{1cm}}@{}}
        \toprule
        \multirow{3}{*}{\textbf{Layer}} & \multirow{3}{*}{\textbf{Score}} & \multicolumn{2}{c}{SCD Datasets} & \multicolumn{2}{c}{ChangeVPR} \\
        \cmidrule(lr){3-4} \cmidrule(lr){5-6}
        && F1-score & mIoU & F1-score & mIoU \\
        \midrule
        Initial& \multirow{2.5}{*}{0.88}  & 61.1 & 43.5 & 61.2 & 42.4 \\
        Intermed. &  & 67.3 & 51.0 & 63.8 & 46.0 \\
        \midrule
        Last & 0.72 & 65.5 & 47.9 & 63.0 & 44.7 \\
        Last & 0.95 & 66.7 & 50.0 & 63.3 & 45.2 \\
        \rowcolor{LightCyan}
        Last & 0.88 & \textbf{67.7} & \textbf{51.3} & \textbf{64.1} & \textbf{46.3} \\
        \bottomrule
    \end{tabular}
    \caption{\textbf{Performance comparison of GeSCF using different mask embedding layers and change confidence score in semantic similarity matching.} }
\label{sup:mask_quan}
\end{table}

\begin{table}[t]
    \footnotesize
    \centering
    \setlength{\tabcolsep}{0.5pt}
    \renewcommand{\arraystretch}{0.3} 
      \begin{tabular}{@{}>{\raggedright}m{1.3cm} @{}>{\centering\arraybackslash}m{1.5cm} @{}>{\centering\arraybackslash}m{1.3cm} *{4}{>{\centering\arraybackslash}m{0.8cm}}@{} @{}>{\centering\arraybackslash}m{0.8cm}}
            \toprule
            \multirow{3}{*}{\textbf{Method}} & \multirow{3}{*}{\textbf{Training}} & \multirow{3}{*}{\textbf{Resolu.}} & \multicolumn{4}{c}{\textbf{Perturbation Scale}} & \multirow{3}{*}{\textbf{Avg.}} \\
            \cmidrule(lr){4-7}
            &  &&\textbf{0\%} & \textbf{5\%} & \textbf{15\%} & \textbf{25\%}  \\
            \midrule
            CSCDNet     &\multirow{6.5}{*}{in-domain}&\multirow{6.5}{*}{256$\times$256} & 71.9 & 70.5 & 70.3 & 69.7 & 70.6 \\
            CDResNet     && &70.1 & 69.2 & 68.6 & 67.9 & 69.0 \\
            DR-TANet     && &70.1 & 69.4 & 68.3 & 67.0 & 68.7 \\
            C-3PO       && &72.8 & 72.2 & 71.9 & 71.1 & 72.0 \\
            \midrule
            \rowcolor{LightCyan}
            \textbf{GeSCF}         &zero-shot& 256$\times$256 & \textbf{74.9} & \textbf{74.2} & \textbf{72.8} & \textbf{71.4} &  \textbf{73.3} \\
            \bottomrule
      \end{tabular}
\caption{\textbf{Quantitative comparison (F1-score) on VL-CMU-CD.}}
\label{perspective}
\end{table}

\begin{table}[t]
    \footnotesize
    \centering
    \setlength{\tabcolsep}{6.2pt}
    \renewcommand{\arraystretch}{0.2} 
      \begin{tabular}{@{}>{\raggedright}m{2cm} @{}>{\centering\arraybackslash}m{1.2cm} *{4}{>{\centering\arraybackslash}m{0.7cm}}@{}}
            \toprule
            \multirow{3}{*}{\textbf{Method}}&  \multirow{3}{*}{\textbf{Training}} & \multicolumn{3}{c}{\textbf{CDnet 2014}} & \multirow{3}{*}{\textbf{Avg.}} \\
            \cmidrule(lr){3-5}
            & & Bus & Tram & Boats \\
            \midrule
            3DCD \cite{Mandal20203DCDSI}  & in-domain  & 0.79 & 0.75 & \textbf{0.88} & 0.81 \\
            \rowcolor{LightCyan}
            \textbf{GeSCF}  & zero-shot  & \textbf{0.80} & \textbf{0.81} & 0.86 & \textbf{0.82} \\
            \bottomrule
      \end{tabular}
      \vspace{-2mm}
\caption{\textbf{Quantitative comparison (F1-score) on CDnet 2014.}}
\label{video}
\end{table}

\begin{figure}[t]
    \centering
    \includegraphics[width=\linewidth]{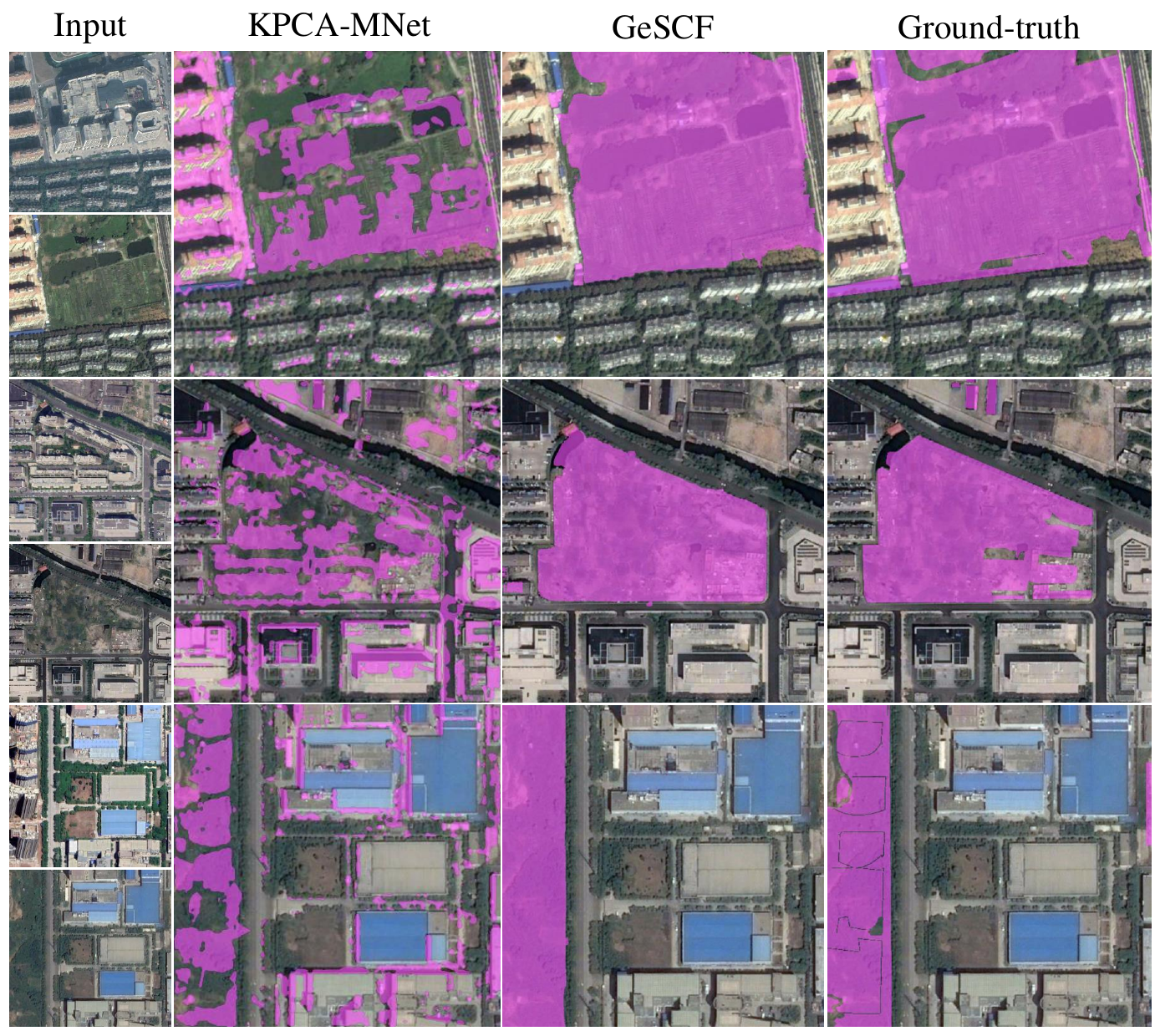}
    \caption{\textbf{Qualitative results of GeSCF on the SECOND (\texttt{test}) benchmark.} Our GeSCF can also perform zero-shot remote sensing CD, demonstrating its versatility and potential across different CD domains.}
    \label{sup:remote}
\end{figure}

\noindent\textbf{Different ViT Backbones.} We incorporate various ViT backbones of different sizes from SAM and DINOv2 \cite{Oquab2023DINOv2LR} (see \cref{sup:backbone}). The results show that ViTs of SAM variants outperform their DINOv2 counterparts and larger ViT yields superior performance compared to smaller models.

\noindent\textbf{Mask Embeddings from the Last Layer.} Previous studies \cite{Ghiasi2022WhatDV} have empirically demonstrated the ViT's transition from low-level feature encoding in early layers to capturing more global, semantic representations in deeper layers. We observe that this hierarchical feature specialization also exists in SAM ViT (see \cref{sup:mask}). Additionally, we perform quantitative ablation studies regarding mask embedding layers and change confidence scores (see \cref{sup:mask_quan}). Our results show that utilizing mask embeddings from the final layer yields superior performance compared to embeddings from the initial or intermediate layers. Furthermore, a high change confidence score can result in the inclusion of irrelevant masks, while a low score becomes overly strict, hindering the detection of actual changes. Therefore, our change confidence score acts as a semantic lower bound within SAM's latent space for accurate change detection.

\subsection{Regarding Registration and Resolution}
To assess robustness against perspective and resolution variations, we simulate perspective distortions by applying random perturbations of varying scales (up to a 64-pixel offset) to images downscaled to half of their original $512\times512$ resolution on the VL-CMU-CD dataset (see \cref{perspective}). While all baselines are trained on full-resolution images, GeSCF maintains a zero-shot setting, augmented with a simple SuperPoint \cite{DeTone2017SuperPointSI} + RANSAC \cite{Fischler1981RandomSC} registration module.

\subsection{Beyond Scene Change Detection}
Although our research primarily focuses on natural scene CD, with each CD domain typically focusing on its specialized area rather than integrating other CD fields \cite{Lee2024SemiSupervisedSC}, we have discovered that our framework can also work as a zero-shot remote sensing CD framework (see \cref{sup:remote} for qualitative examples). Despite SAM being exclusively trained on natural images from photographers, SAM's feature space and our proposed feature strategies remain effective when applied to remote sensing data. Furthermore, we also evaluate our proposed GeSCF against the video sequence CD method from \cite{Mandal20203DCDSI} on the CDnet 2014 \cite{Wang2014CDnet2A} benchmark (see \cref{video}). The experimental results further confirm the exceptional generalizability of GeSCF. Although these results are preliminary, they indicate that the representations from SAM may be useful for different CD domains.

\section{Training Objectives}
To comprehensively evaluate temporal consistency, we train all baselines using both uni-temporal and bi-temporal training objectives \cite{Zheng2021ChangeIE}.

\noindent\textbf{Uni-temporal Objective} is a standard binary cross-entropy loss, a widely adopted loss function for binary classification tasks. It quantifies the discrepancy between the predicted change probabilities and the ground-truth, effectively guiding the network to accurately distinguish between change and no-change regions. The uni-temporal objective is calculated as follows:

\begin{equation}
    \mathcal{L}_{bce} = -[y\log(p) + (1 - y)\log(1 - p)], \\ 
\end{equation}

\noindent where $y\in{0, 1}$ specifies the ground-truth class and $p \in [0, 1]$ denotes predicted probability for positive class. 

\noindent\textbf{Bi-temporal Objective} \cite{Zheng2021ChangeIE} is specifically designed for binary change detection by leveraging paired images captured at different time steps. By jointly optimizing with both temporal directions, it complements the uni-temporal objective and further refines the model's temporal consistency. The bi-temporal objective is formulated as follows:

\begin{equation}
    \mathcal{L}_{bce}^{t0 \leftrightarrow t1} = m \mathcal{L}_{bce}^{t0 \rightarrow t1} + n \mathcal{L}_{bce}^{t1 \rightarrow t0}, \\  
\end{equation}

\noindent where $m$ and $n$ are set to $0.5$ and $t0$$\rightarrow$$t1$ (or $t1$$\rightarrow$$t0$) represents the concatenation order of the input images. Note that $\mathcal{L}_{bce}^{t0 \leftrightarrow t1}$ is equal to $\mathcal{L}_{bce}^{t0 \rightarrow t1}$ (or $\mathcal{L}_{bce}^{t1 \rightarrow t0}$) if the model entails temporally symmetric architectures.

\section{Limitations}
Generalizable scene change detection is an emerging and challenging task essential for advancing the SCD community. We are the first to define \textbf{\textit{what is generalizable scene change detection}} and to introduce a straightforward yet powerful framework, complemented by a broader domain evaluation dataset and a comprehensive evaluation protocol. However, certain limitations persist, including the need for demonstration across more diverse domains, such as various indoor environments, and challenges related to the bias of SAM. These limitations leave room for subsequent research and interdisciplinary studies \cite{Ke2023SegmentAI, Chen2024RobustSAMSA} to enhance the robustness of our framework further.

\clearpage
\newpage
{
    \small
    \bibliographystyle{ieeenat_fullname}
    \bibliography{supplementary}
}